\documentclass[10pt,twocolumn,letterpaper]{article}

\usepackage{cvpr}              

\usepackage{multirow}
\usepackage{tabularx}
\newcolumntype{C}{>{\centering\arraybackslash}X}

\usepackage{pgfplots}
\pgfplotsset{compat=1.5, every axis/.append style={font=\small, /pgf/number format/1000 sep={}}}
\usepackage{tikz}
\usetikzlibrary{quotes, arrows.meta, angles, calc, 3d, shapes, intersections, plotmarks}
\usepgfplotslibrary{statistics, polar, groupplots}
\pgfplotsset{every mark/.append style={solid},} 
\usepackage{tikzscale}

\usepackage[ruled, linesnumbered, noend]{algorithm2e}
\SetKwComment{Comment}{\% }{}

\usepackage{amsmath,amsfonts,bm}

\def\1{\bm{1}}

\def\vx{{\bm{x}}}

\DeclareMathAlphabet{\mathsfit}{\encodingdefault}{\sfdefault}{m}{sl}
\SetMathAlphabet{\mathsfit}{bold}{\encodingdefault}{\sfdefault}{bx}{n}

\def\gI{{\mathcal{I}}}

\newcommand{\R}{\mathbb{R}}

\newcommand{\reals}{\mathbb{R}}
\newcommand{\transpose}{^\mathsf{T}}
\newcommand{\gd}{\dot{\gamma}}
\newcommand{\gdd}{\ddot{\gamma}}
\newcommand{\gh}{\hat{\gamma}}
\newcommand{\gdh}{\hat{\dot{\gamma}}}
\newcommand{\dd}[2]{\frac{\text{d} #1}{\text{d} #2}}
\newcommand{\ddt}[1]{\frac{\text{d} #1}{\text{d}t}}
\newcommand{\dt}{\text{ d}t}

\newcommand{\pp}[2]{\frac{\partial #1}{\partial #2}}
\newcommand{\ddelta}[2]{\frac{\delta #1}{\delta #2}}

\usepackage{xspace}
\makeatletter
\DeclareRobustCommand\onedot{\futurelet\@let@token\@onedot}
\def\@onedot{\ifx\@let@token.\else.\null\fi\xspace}
\def\wrt{w.r.t\onedot}

\def\viz{viz\onedot}
\def\eg{e.g\onedot} 
\def\ie{i.e\onedot} 
 
\def\etc{etc\onedot}

\makeatletter
\renewcommand\paragraph{
  \@startsection{paragraph}
                {4}
                {\z@}
                {1.0ex \@plus0.5ex \@minus.2ex}
                {-0.5em}
                {\normalfont\normalsize\bfseries}}
\makeatother

\usepackage{amssymb}
\usepackage{pifont}
\newcommand{\xmark}{\ding{55}}%
\usepackage{makecell} 
\usepackage{tabularx}
\usepackage{booktabs}

\definecolor{cvprblue}{rgb}{0.21,0.49,0.74}
\usepackage[pagebackref,breaklinks,colorlinks,allcolors=cvprblue]{hyperref}

\def\paperID{13954} 
\def\confName{CVPR}
\def\confYear{2025}

\title{Probability Density Geodesics in Image Diffusion Latent Space}

\author{
Qingtao Yu$^{1}$ , 
\quad Jaskirat Singh$^{1}$ ,  
\quad Zhaoyuan Yang$^{2}$ , 
\quad Peter Henry Tu$^{2}$ ,  \\
\quad Jing Zhang$^{1}$ , 
\quad Hongdong Li$^{1}$ , 
\quad Richard Hartley$^{1}$ , 
\quad Dylan Campbell$^{1}$  \\
$^{1}$Australian National University \quad $^{2}$GE Research
\\{\tt\small \{terry.yu,jaskirat.singh,jing.zhang,hongdong.li,richard.hartley,dylan.campbell\}@anu.edu.au} \\
{\tt\small \{peter.tu,zhaoyuan.yang\}@ge.com}
}

\begin{document}
\maketitle

\begin{abstract}

Diffusion models indirectly estimate the probability density over a data space, which can be used to study its structure. In this work, we show that geodesics can be computed in diffusion latent space, where the norm induced by the spatially-varying inner product is inversely proportional to the probability density. In this formulation, a path that traverses a high density (that is, probable) region of image latent space is shorter than the equivalent path through a low density region. We present algorithms for solving the associated initial and boundary value problems and show how to compute the probability density along the path and the geodesic distance between two points. Using these techniques, we analyze how closely video clips approximate geodesics in a pre-trained image diffusion space. Finally, we demonstrate how these techniques can be applied to training-free image sequence interpolation and extrapolation, given a pre-trained image diffusion model. Code is availiable at \href{https://github.com/TerrysLearning/GeodesicDiffusion}{https://github.com/TerrysLearning/GeodesicDiffusion}
\end{abstract}


\section{Introduction}
\label{sec:intro}

\begin{figure}[!t]\centering
\begin{subfigure}[]{0.4\linewidth}\centering
\includegraphics[width=\linewidth, trim=10pt 10pt 20pt 20pt, clip]{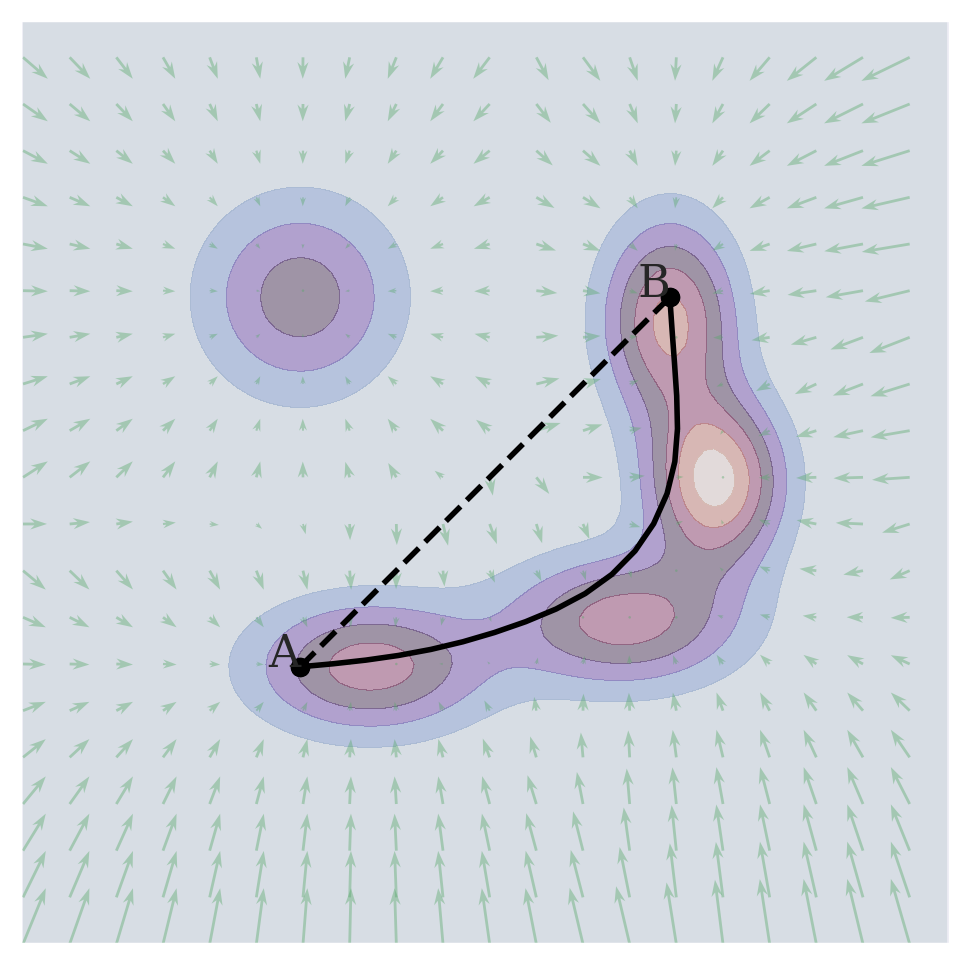}%
\caption{}%
\label{fig:splash_geodesic}
\end{subfigure}\hfill
\begin{subfigure}[]{0.58\linewidth}\centering
\includegraphics[width=\linewidth, trim=5pt 20pt 10pt 10pt, clip]{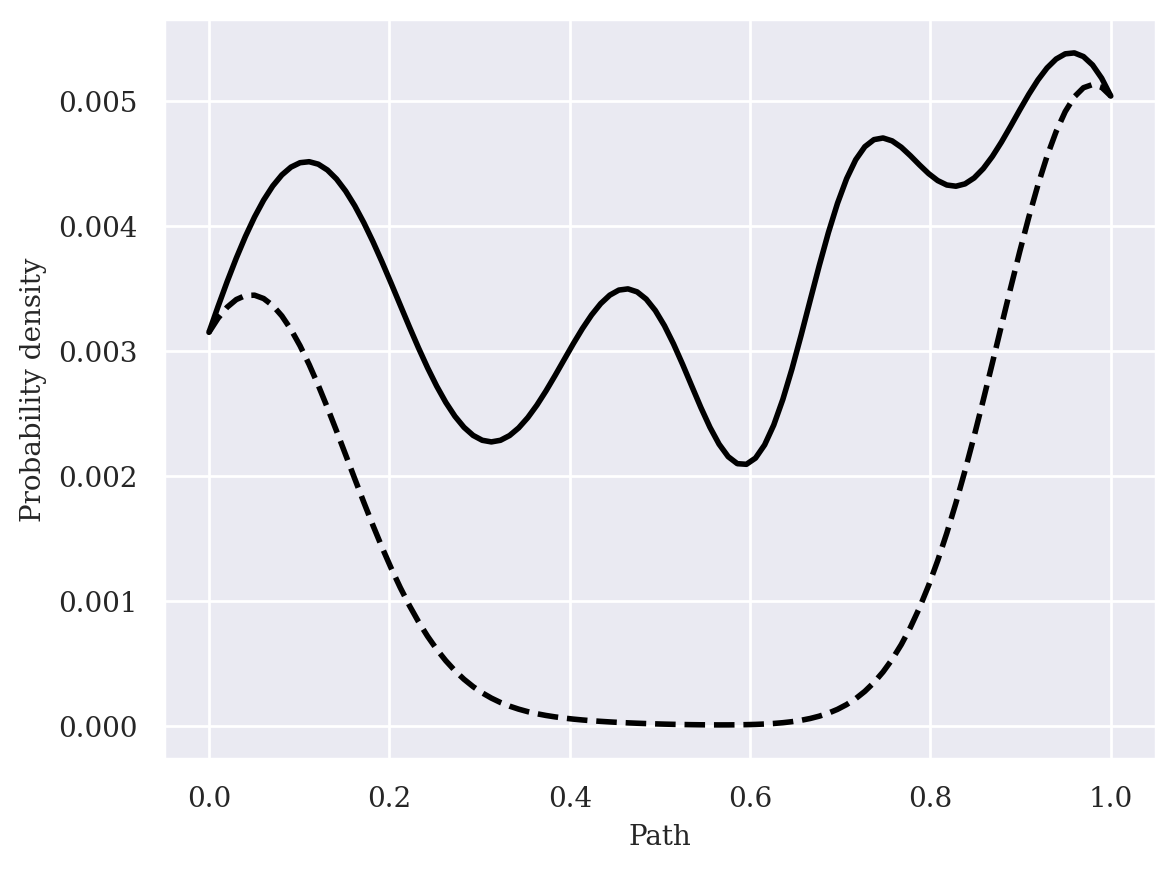}%
\caption{}%
\label{fig:splash_probability}
\end{subfigure}\vfill
\begin{subfigure}[]{\linewidth}\centering
\includegraphics[width=\linewidth, trim=0pt 0pt 0pt 0pt, clip]{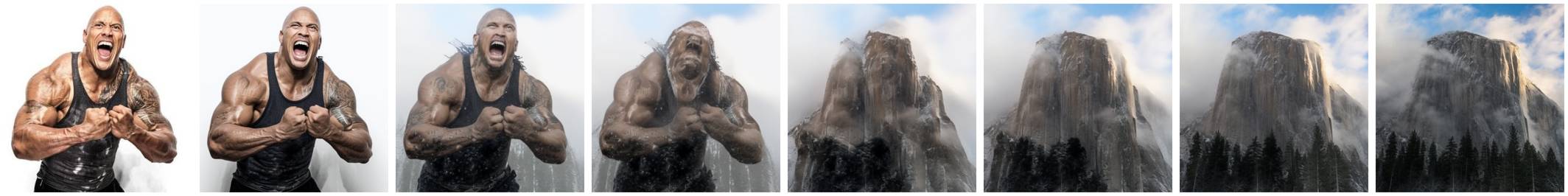}
\caption{}%
\label{fig:splash_images}
\end{subfigure}%
\caption{
Given a probability density and initial or boundary conditions (here, the position of two points A and B), geodesics can be computed in this space.
If the norm is chosen to be inversely proportional to the probability density, these geodesics preferentially traverse high density regions of the space.
For an image data space, these correspond to plausible, realistic images according to the probability density, such as that learned by an image diffusion model.
Here, we show the outputs of our boundary value problem (BVP) solver that computes a geodesic between endpoints A and B on a toy 2D example.
\subref{fig:splash_geodesic}~The geodesic and straight-line trajectories between A and B, given the underlying visualized probability density field (contours) and its gradient (arrows).
\subref{fig:splash_probability}~Probability density curves for both trajectories, showing that the straight-line path drops to zero probability very rapidly whereas the geodesic remains in higher probability regions.
\subref{fig:splash_images}~Images corresponding to points along a geodesic in Stable Diffusion \cite{rombach2022high} latent space, given the left and right endpoints, computed using our BVP solver.
}
\label{fig:splash}
\end{figure}

When trained over a data space, a diffusion model \cite{sohl2015deep,song2019generative,song2020denoising,song2021score} can tell us the direction in which a data point of that space should move in order to increase its likelihood. 
That is, it learns a vector field corresponding to the gradient of the log-probability of the data (the Stein score function \cite{stein1972bound}).
This is sufficient information to define a Riemannian manifold in this space, where the norm induced by a spatially-varying inner product is chosen to be inversely proportional to the probability density \cite{sorrenson2024learning, trillos2024fermat, little2022balancing}.
This has the effect that geodesics---loosely, shortest paths---on this manifold `prefer' to traverse high-density regions, which can be thought of as shortcuts in the space.
In the context of image latent diffusion models \cite{rombach2022high}, these regions correspond to the latent vectors of probable images, while low-density regions correspond to unrealistic image latents.
By computing geodesics in this space, we can find shortest paths between images and study the structure of the learned space.

In this work, we outline the relevant theory needed for computing probability density geodesics in diffusion latent space and present algorithms for solving the associated initial and boundary value problems.
A 2D toy example of the boundary value problem (BVP) is presented in \cref{fig:splash}, alongside image outputs from the BVP solved in the 16384-D latent space of Stable Diffusion \cite{rombach2022high}.
We address several challenges associated with computing geodesics in this space:
(i)~accurately estimating the score function, especially in low-density regions;
(ii)~handling (potentially spatially-varying) conditional probability densities; and
(iii)~solving initial and boundary value problems efficiently in an extremely high-dimensional space.

We also describe how to compute several useful quantities for analysis: the relative probability density along the path, the geodesic distance between two points, and the geodesic gradient norm.
Using these techniques, we analyze how closely video clips approximate geodesics in a pre-trained image diffusion space. 
This work is part exploratory (\eg, what image sequences correspond to geodesics?) and part applied (\eg, how can these techniques be used to solve image sequence-based tasks?).
For the latter, we evaluate the performance of these techniques on training-free image sequence interpolation and extrapolation tasks, given a pre-trained image diffusion model. Our contributions are:
\begin{enumerate}
    \item characterizing probability density geodesics in diffusion latent space;
    \item algorithms for solving initial and boundary value problems that address challenges specific to conditional diffusion models; and
    \item an application to image sequence interpolation and extrapolation tasks.
\end{enumerate}
We evaluate our training-free approach on five datasets with respect to other interpolation methods, including those that fine-tune the underlying diffusion model, achieving state-of-the-art results.

\section{Related Work}
\label{sec:relwork}

\paragraph{Text-to-image diffusion models.}
Recently, diffusion models \cite{ho2020denoising,dhariwal2021diffusion} have gained significant attention in deep generative modeling.
Given specific input conditions, these models generate realistic data by progressively denoising Gaussian noise through a noise prediction network, resulting in outputs that closely resemble real data distributions based on the provided conditions.
With the availability of large-scale image-text pairs for training, diffusion models have achieved remarkable performance in the text-to-image generation task \cite{saharia2022photorealistic,rombach2022high,podell2024sdxl}.
Due to the flexibility of textual prompting, text-to-image diffusion models have been widely applied to image editing \cite{Brooks_2023_CVPR,hertz2023prompttoprompt,DBLP:conf/cvpr/KawarZLTCDMI23,DBLP:conf/iccv/HertzAC23}, personalization \cite{DBLP:conf/cvpr/RuizLJPRA23,DBLP:conf/cvpr/KumariZ0SZ23,gal2023an,DBLP:conf/iccv/HanLZMMY23}, stylization \cite{DBLP:conf/cvpr/TumanyanGBD23,DBLP:journals/corr/abs-2303-09522,DBLP:conf/cvpr/HertzVFC24}, \etc.
In this work, we use a pretrained Stable Diffusion \cite{rombach2022high} model
with which to explore probability density geodesics for video analysis, image interpolation and extrapolation.

\paragraph{Image interpolation.}
Image interpolation, also known as image morphing, refers to the process of generating a smooth transition between two images by creating intermediate images \cite{wolberg1998image,zope2017survey}.
Traditional methods rely on pixel-based transformations, such as mesh morphing or defining a blending function \cite{Image_Metamorphosis_LeeWCS96,seitz1996view,zhu2007image}. 
However, these methods typically require pre-labeled corresponding points and human involvement.
Some attempts have been made to automate this process \cite{pim_morphing,Automating_Image_Morphing_ACM_2014}, but they generally work only on image pairs generated in controlled environments. 
With advances in generative models, Generative Adversarial Networks (GANs) \cite{NIPS2014_5ca3e9b1} have been applied to the image interpolation task \cite{DBLP:journals/tog/ParkSN20,kim2022polymorphic,DBLP:conf/cvpr/SimonA20} and achieve moderate success.
However, GAN-based image interpolation frameworks do not perform well in open-world settings.
Specifically, input image pairs are required to be in-distribution with the training data of GAN models, which limits the application scenarios for image interpolation.
With recent progress in text-to-image diffusion models, several works \cite{zhang2024diffmorpher,yang2024impus,wang2023interpolating,guo2024smooth,he2024aid} have shown that these models achieve impressive results in image interpolation.
Given any image pair, through textual embedding inversion and embedding interpolation, relatively smooth intermediate images can be obtained.
In this work, we also employ a diffusion model; however, compared with prior works \cite{zhang2024diffmorpher,yang2024impus,guo2024smooth}, our method is training-free and can be applied to image extrapolation. 
Furthermore, our approach offers a theoretical framework with which to understand these tasks.

\section{Probability Density Geodesics}
\label{sec:theory}

In this section, we will outline the theory of probability density geodesics, followed by the algorithmic development.
The following section will apply the theory to a diffusion latent space and address some design decisions particular to that context.
We aim to show that the structure of a data space can be studied by computing geodesics with respect to the probability distribution of the data, which can be obtained by training a diffusion model in that space.

A geodesic is the (locally) shortest path between two points on a Riemannian manifold 
with a constant speed parameterization
and can be obtained by minimizing the length of a curve between them using the calculus of variations.
In the following, time derivatives are denoted $\dot{x}$ and second derivatives as $\ddot{x}$.

\paragraph{Equation for path length.}
Let $\gamma : [0, 1] \rightarrow \R^n$ be a path such that $\gamma(0) = x_0$ and $\gamma(1) = x_1$, and $S : \{\gamma_i\} \rightarrow \R$ be the action functional on the set of such paths, defined by
\begin{align}
S[\gamma] &= \int_a^b L(t, \gamma(t), \gd(t)) \dt,
\end{align}
where $L$ is the Lagrangian given by
\begin{align}
L(t, \gamma, \gd) &= \sqrt{ \langle \gd, \gd \rangle_{\!K(\gamma)}} \text{ with } K(\gamma) = p(\gamma)^{-2} I.
\end{align}
Then $S[\gamma]$ is the path length for path $\gamma$.
The norm $\|x\| = \sqrt{ \langle x, x \rangle_{\!M}}$ induced by the inner product $\langle x, y \rangle_M = x\transpose M y$ is inversely proportional to the probability density with function $p : \R^n \rightarrow \R_+$.
The interpretation here is that paths that pass through high density regions are `shorter' than paths through low density regions.
Note that this reduces to the standard formulation for the path length of a curve when the probability distribution is uniform.

\paragraph{Euler--Lagrange equations.}

Given this definition, a path $\gamma$ is a stationary point of $S$ if and only if it satisfies the Euler--Lagrange equations, \viz,
\begin{align}
\pp{}{\gamma} L(t, \gamma(t), \gd(t)) - \ddt{}\pp{}{\gd} L(t, \gamma(t), \gd(t)) &= 0.
\end{align}
Using $\nabla \log p(\gamma) = \frac{1}{p} \dd{p}{\gamma}\transpose$ and a constant speed parameterization of the path, for our Lagrangian we obtain 
\begin{align}
\gdd + \|\gd\|^2 \left( I - \gdh\gdh\transpose \right) \nabla \log p(\gamma) &= 0,
\label{eq:ode}
\end{align}
where the unit velocity is given by $\gdh = \gd / \|\gd\|$.
That is, we obtain a nonlinear second-order ordinary differential equation (ODE) that expresses the relationship between the scaled acceleration and the gradient of the log probability.

It is important to observe here that $\nabla \log p(\gamma)$ is the Stein score function \cite{stein1972bound,hyvarinen2005estimation}, the exact quantity that is estimated by a score-based diffusion model.
The full derivation is presented in the appendix.

\paragraph{Functional derivative.}

This second-order ODE expresses the relationship at optimality, \ie, given an initial position and velocity we can obtain the associated optimal path. However, we can also derive the functional derivative $\ddelta{S}{\gamma}$ of the path length functional $S$ by approximating the curve by a polygonal line with $n$ segments, as $n$ grows arbitrarily large. We obtain, for any $\gamma$
with a constant speed parameterization,
\begin{align}
\ddelta{S}{\gamma} &= \frac{-1}{p(\gamma)\|\gd\|} \left( \left( I - \gdh\gdh\transpose \right) \nabla \log p(\gamma) + \frac{\gdd}{\|\gd\|^2} \right).
\label{eq:derivative}
\end{align}

\paragraph{Probability density along the path.}

While estimating the absolute probability of any data point is challenging, the probability relative to the starting point is convenient to compute.
For a conservative vector field $\nabla \log p$ (the gradient of a scalar field), line integrals are path independent.\footnote{This assumption is not strictly true for standard diffusion models, but is a good approximation \cite{chao2023investigating}.}
Therefore, the relative log-probability between two points $\gamma(a)$ and $\gamma(t)$, along \textit{any} path $\gamma$ connecting them, is
\begin{align}
\log \tilde{p}_a(\gamma(b)) &:=
\log p(\gamma(b) - \log p(\gamma(a))\\
&= \int_a^b  \gd(t)\transpose \nabla \log p(\gamma(t)) \dt .
\label{eq:log_probability}
\end{align}
Let $f(x) = \gd(x)\transpose \nabla \log p(\gamma(x))$ and let $t_i = a+i/n$ for $0 \leq i \leq n$ be $n+1$ equally spaced samples along the path.
Then by the trapezoidal rule, the relative probability at these points along the path can be approximated as
\begin{align}
\log \tilde{p}_a(\gamma(t_i)) &\approx \! \frac{1}{n} \left(\! \frac{f(a)}{2}  \!+\! \sum_{k=1}^{i-1} f(t_k) \!+\! \frac{f(t_i)}{2}  \!\right) .
\end{align}
Higher-order approaches, such as Simpson's rule, may also be used for this approximation.
From this, we may compute the relative probability $\tilde{p}_a(\gamma(t_i)) = p(\gamma(t_i)) / p(\gamma(a))$ at every point $\gamma(t_i)$ along the curve.

\paragraph{Geodesic distance.}
Given the probabilities computed in the previous section, it is trivial to estimate the geodesic distance.
Let $t_i = a+i/n$ for $0 \leq i \leq n$ be $n+1$ equally spaced samples along the geodesic $\gamma$, then the relative geodesic distance is given by

\begin{align}
\label{eq:geo_distance}
\tilde{d}_a(b) &=
\int_a^b  \frac{\|\gd(t)\|}{\tilde{p}_a(\gamma(t))} \dt \\
&\approx \! \frac{1}{n} \!\left(\! \frac{\|\gd(a)\|}{2} \!+\! \sum_{i=1}^{n-1}\! \frac{\|\gd(t_i)\|}{\tilde{p}_a(\gamma(t_i))} \!+\! \frac{\|\gd(b)\|}{2\tilde{p}_a(\gamma(t_n))} \!\right) \nonumber
\end{align}
and the absolute geodesic distance is given by $d(a,b) = \tilde{d}_a(b) / p(\gamma(a))$ .

\section{Geodesics in Diffusion Latent Space}
\label{sec:method}
In this section, we show how to compute geodesics in the latent space of a pre-trained Stable Diffusion \cite{rombach2022high} model.
An element $x \in \reals^{64 \times 64 \times 4}$ of this latent space can be obtained from an image $\gI$ via an encoder $\mathcal{E}(\gI)$; a decoder $\mathcal{D}(x)$ performs the inverse mapping.
A forward process incrementally corrupts these elements with Gaussian noise, up to timestep $\tau=T$, resulting in a noise element $x_T \sim \mathcal{N}(0, I)$.
To distinguish between the time dependence of the diffusion process and that of the path formulation in \cref{sec:theory}, we use $\tau$ instead of $t$ to denote the diffusion time-step throughout the rest of the paper.
The diffusion model is trained on a very large dataset of images to denoise elements of this latent space at all timesteps.
It is parameterized by a neural network $\epsilon_\theta (x_{\tau};z)$ that predicts the noise $\epsilon$ that was used to produce noisy sample $x_{\tau}$ from clean sample $x_0$ \cite{ho2020denoising}.
This is proportional to the score function $\nabla_{x_{\tau}} \log p_{\tau} (x_{\tau})$ of the smoothed (noised) density $p_{\tau}$ \cite{ho2020denoising}.
Finally, our method makes use of the deterministic DDIM forward and backward processes (DDIM-F, DDIM-B) with noise schedule parameter $\overline{\alpha}_{\tau}$, known as inversion \cite{song2020denoising}, defined by
\begin{align}
\dd{}{\tau} \left( \frac{x_{\tau}}{\sqrt{\overline{\alpha}_{\tau}}} \right)
=
\dd{}{\tau} \left(\sqrt{\frac{1-\overline{\alpha}_{\tau}}{\overline{\alpha}_{\tau}}} \right) \epsilon_\theta\left(x_{\tau};z \right).
\label{eq:ddim_inversion}
\end{align}

Working with diffusion models introduces several challenges, which we address in this section.
In particular,
(i) how to handle (potentially varying) conditional probability densities;
(ii) how to accurately estimate the score function $\nabla \log p(x)$, especially in low density regions; and
(iii) how to solve the initial and boundary value problems efficiently in an extremely high-dimensional space (\eg, $\reals^{16384}$). 

\subsection{Conditional probability density geodesics}
\label{sec:method_cond}

Most image diffusion models are conditional models, trained to denoise an image conditioned on a signal $z$ such as a CLIP-encoded text prompt \cite{radford2021learning, rombach2022high, saharia2022photorealistic}.
Hence, we extend the formulation from \cref{sec:theory} to consider the conditional probability density function $p(\gamma(t) \mid \zeta(t, \gamma(t))$.
Here, $\zeta : \reals \times \reals^n \rightarrow \reals^d$ is a function over the path parameter and latent space that obtains the values of the conditioning vectors $z_a$ and $z_b$ at the endpoints, such that $\zeta(0, \gamma(0)) = z_0$ and $\zeta(1, \gamma(1)) = z_1$ for $(a,b)=(0,1)$.
In our experiments, we primarily consider
the case where $\zeta$ is linear in time and constant in the latent space: $\zeta(t) = (1 - t) z_0 + t z_1$.

\subsection{Score estimation}
\label{sec:method_score}
The score function is likely to be poorly estimated in low-density regions \cite{song2019generative}.
This occurs since the diffusion model does not have enough evidence during training to estimate the score accurately in regions of the latent space where data is scarce.
However, estimating accurate directions in these regions is critical for solving boundary value problems, since the initial path to be optimized is likely to traverse low-density regions.
Diffusion models apply multiple levels of noise to data elements in order to alleviate this problem.
By doing so, low-density regions of the space can be populated by noised data, improving the quality of the estimated gradients at those locations.
As a result, we choose to compute geodesics (and query the diffusion model) at a particular non-zero noise level, with the associated diffusion timestep $\tau$.
A side benefit of this is that the probability density becomes increasingly Gaussian as the timestep increases, which provides a useful inductive bias, as explained in the next section.

However, there is an additional complexity.
For a noised training sample $x_{\tau}$, the unconditional model $\epsilon_\theta (x_{\tau}; z=\varnothing, \tau)$ is expected to predict the noise $\epsilon$ that was used to produce this sample from clean sample $x_0$.
Unfortunately, in our case, we obtain vectors $x_{\tau}$ from an optimization procedure, initialized along a great circle between endpoints $x_{a,\tau}$ and $x_{b,\tau}$.
These points are (at least initially) out-of-distribution (OOD) for the model, and we should not expect it to predict the noise well.
Instead, \citet{katzir2024noise} observe that the prediction consists of a domain correction term $\delta_\text{D}$ and a denoising term $\delta_\text{N}$.
The residual $\delta_\text{N} - \epsilon$, as used in score distillation sampling \cite{poole2023dreamfusion}, is generally non-zero and noisy, leading to an averaging (over-smoothing) effect in the optimized latents.
In our optimization procedure, we are not aiming to denoise the samples, so neglect the denoising direction $\delta_\text{N}$ and instead use noise-free score distillation~\cite{katzir2024noise} direction $\phi (x \mid z, \tau)$ to approximate the gradient direction,
\begin{align}
&\nabla\log p(x \mid z, \tau) \approx \beta \phi(x \mid z, \tau) \label{eq:nabla_logp}\\
&\phi(x \!\mid\! z, \tau) \!=\! \mathbb{E}_{\tau'\!\in \mathcal{R}_{\tau}} w(\!\tau') \left( \sigma d(x_{\tau'\!} \!\mid\! z) \!- \! d(x_{\tau'\!} \!\mid\! z_\text{neg}) \right),
\label{eq:score_distillation}
\end{align}
where $\beta$ is a scalar hyperparameter,
$w(\tau)$ is a weighting function,
$\sigma$ is the classifier-free guidance parameter (0 if unconditional) \cite{ho2021classifier},
$z_\text{neg}$ is a negative prompt embedding that says something about the specific OOD latent vector (see \cref{sec:app_hyperparameters}),
$d(x_{\tau}|z) = \epsilon_\theta(x_{\tau}|\varnothing) - \epsilon_\theta(x_{\tau}|z)$
is the direction function, and
the expectation is taken over a range of timesteps $\mathcal{R}_{\tau} = [\tau-\Delta\tau, \tau + \Delta\tau]$.

\subsection{Efficient IVP and BVP solvers}
\label{sec:method_alg}
\begin{figure}[!t]\centering
     \begin{subfigure}[]{\linewidth}\centering 
        \includegraphics[width=\linewidth]{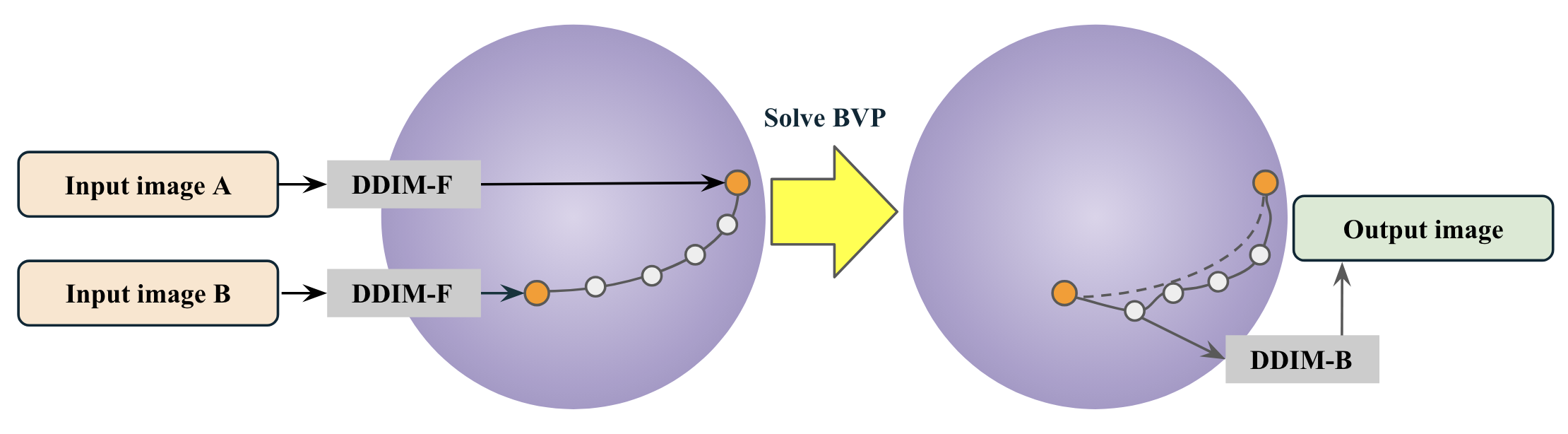}
        \caption{The pipeline for image interpolation.}
        \label{fig:bvp-pipeline}
    \end{subfigure}\vfill
    \begin{subfigure}[]{\linewidth}\centering 
        \includegraphics[width=\linewidth]{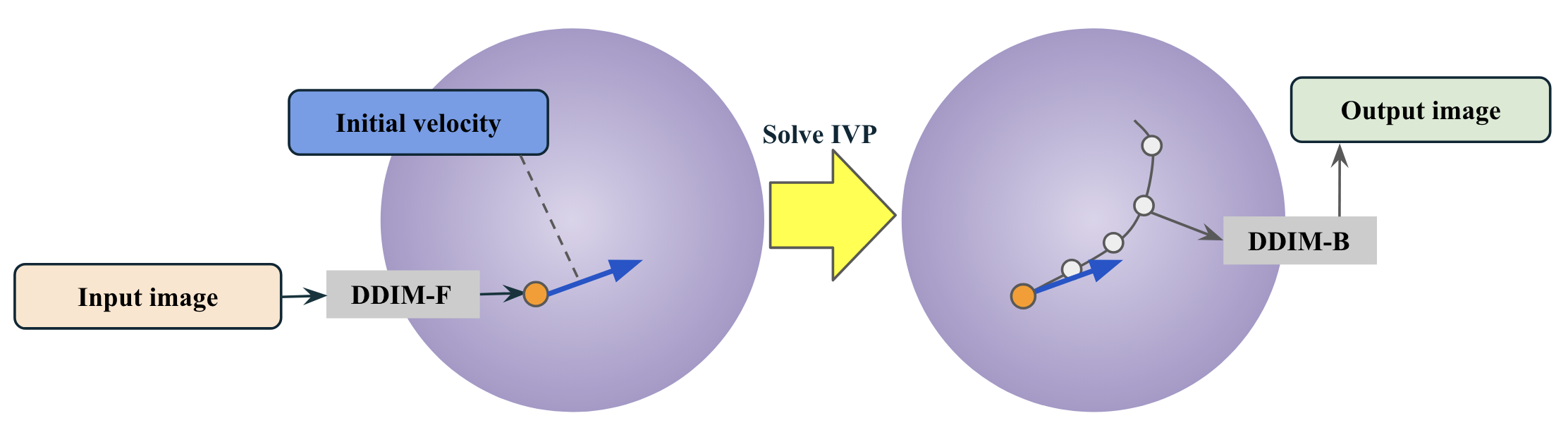}
        \caption{The pipeline for image extrapolation.}
        \label{fig:ivp-pipeline}
    \end{subfigure}%
    \caption{Pipelines for the image interpolation and extrapolation tasks, addressed by solving boundary and initial value problems in diffusion latent space.}
    \label{fig:pipelines}
\end{figure}

Solving the initial and boundary value problems efficiently with respect to time and memory becomes challenging in the extremely high-dimensional latent space $\reals^{4\times64\times64}$ of Stable Diffusion \cite{rombach2022high}.
For this dimensionality, standard techniques, such as collocation algorithms for solving BVPs \cite{kierzenka2001bvp}, become prohibitive in time and memory.
We also aim to minimize calls to the diffusion model, since this is expensive to compute.
In view of these aims, we use a simple low-memory parametrization and optimization strategy.

Before outlining the algorithms, we first observe that the probability distribution becomes more Gaussian as the diffusion timestep increases \cite{chen2024varying} and that the Gaussian Annulus Theorem \cite{blum2020foundations} states that a point chosen at random from a unit variance $d$-dimensional Gaussian distribution will be located in a small annulus around a sphere of radius $\sqrt{d}$ with high probability.
However, for finite-sized gradient descent steps, the optimizer may take the curve away from the sphere, moving into low-density regions where probability density gradients are poorly estimated.
To mitigate this, we reproject the curve back onto the sphere, ensuring that optimization remains within the high-density region where gradient estimates are more reliable, and
we project the functional derivative to the tangent space of the sphere,
\begin{align}
g = ( I \!-\! \gh\gh\transpose) \ddelta{S}{\gamma}.
\label{eq:derivative_sphere}
\end{align}

\paragraph{BVP.}
Given a start position $x^{(0)}$ and an end position $x^{(k+1)}$, we parameterize the path $\gamma$ with a set of control points $\{x_{\tau}^{(i)}\}_{i=1}^{k}$ connected by a spherical piecewise linear function (great circle arcs).
For clarity, we notate points on this curve as $x_t$ with the curve parameter $t \in [0,1]$,
dropping $\tau$ and the ordinal superscript.
\cref{alg:bvp} initializes the path as a great circle.
It then computes the projected gradient descent update for the control points using \cref{eq:derivative_sphere}, where the associated velocities $\gd$ and accelerations $\gdd$ are obtained from a natural cubic spline fit to the control points and end points, and projects the updated control points back onto the sphere.
To save computation, the algorithm uses a coarse-to-fine discretization, where the number of control points $k \in [1, 3, 7, 15]$ varies as optimization progresses, using a bisection strategy.

\begin{algorithm}[!t]
\caption{BVP solver for image interpolation.}
\label{alg:bvp}
\KwIn{ 
start/end image $\mathcal{I}_0 / \mathcal{I}_1$; 
text prompts $p_0/p_1$;
diffusion timestep $\tau$; 
learning rate $\eta$; 
optimizer steps $n$; 
hyperparameter $\beta$;
VAE encoder $\mathcal{E}$; 
CLIP text encoder $\mathcal{C}$
}
\KwOut{geodesic $\gamma$}
$\{z_0, z_1\} \gets \{ \mathcal{C}(p_0), \mathcal{C}(p_1) \}$ \\
$\{x_{0}, x_{1} \}\gets \{\text{DDIM-F}( \mathcal{E}(\mathcal{I}_i), \tau, z_i)\}_{i=\{0,1\}}$ \\
$\gamma \gets \text{Interpolate}(\{(0,x_{0}),(1, x_{1})\})$ \\
 \For{$i = 1$ \KwTo $n$}{
 $ \mathcal{T} \gets \text{TimeSampler}(i)$ \\
 $\mathcal{S} \gets \{(0,x_{0}),(1, x_{1})\}$ \\
  \ForAll{$t \in \mathcal{T}$}{
  $x \gets \gamma(t)$ \;
  $z \gets (1 - t) z_0 + t z_1$ \\
  $s \gets \text{Score}(x,z,\tau, \beta)$ \Comment*[r]{\cref{eq:nabla_logp}} 
  $ g \gets \text{FuncDeriv}(s , x, \dot{x}, \ddot{x})$ \Comment*[r]{\cref{eq:derivative_sphere}} 
  $x \gets \|x\| (x - \eta g) / \|x - \eta g\|$ \\
  $\mathcal{S} \gets \mathcal{S} \cup \{ (t, x)\}$ \\
   }
   $\gamma \gets \text{Interpolate}(\mathcal{S})$
 }
\end{algorithm}

\begin{algorithm}[!t]
\caption{IVP solver for image extrapolation.}
\label{alg:ivp}
\KwIn{
image $\mathcal{I}_0$;
source text prompt $p_0$;
target text prompt $p_1$;
diffusion timestep $\tau$;
optimizer steps $n$;
VAE encoder $\mathcal{E}$; 
CLIP text encoder $\mathcal{C}$;
hyperparameter $\beta$
}
\KwOut{image sequence $\gI$; geodesic $\gamma$}
$\{z_0, z_1\} \gets \{ \mathcal{C}(p_0), \mathcal{C}(p_1) \}$ \\
$x \gets \text{DDIM-F}( \mathcal{E}(\mathcal{I}_0), \tau, z_0)$ \\
$ \dot{x} \gets \text{GetInitVelocity}(x, z_0, z_1)$ \\
$\mathcal{S} \gets \{(0,x)\}$ \\
$ \gI \gets \{\gI_0\}$ \\
\For{$i = 1$ \KwTo $n$}{
    $z \gets (1 - i/n) z_0 + (i/n) z_1 $ \\
    $s \gets \text{Score}(x,z,\tau, \beta)$ \Comment*[r]{\cref{eq:nabla_logp}} 
    $ \ddot{x} \gets \text{ODE}(s, x, \dot{x})$ \Comment*[r]{\cref{eq:ivp_acc}} 
    $ \dot{x} \gets \text{RK4}(\dot{x}, \ddot{x}, 1/n) $ \\
    $ x \gets \|x\| \text{RK4}(x, \dot{x}, 1/n) / \|\text{RK4}(x, \dot{x}, 1/n)\|$ \\
    $ \mathcal{S} \gets \mathcal{S} \cup \{(i/n, x)\} $ \\
    $ \gI \gets \gI \cup \{\mathcal{D}(\text{DDIM-B} (x, \tau, z))\}$ 
}
$\gamma \gets \text{Interpolate}(\mathcal{S})$ 
\end{algorithm}

\paragraph{IVP.}
\cref{alg:ivp} takes an image $\mathcal{I}_0$ and its corresponding text description $p_0$, and extrapolates how the geodesic evolves given an initial velocity formulated from the target prompt $p_1$. 
As with \cref{eq:derivative_sphere}, the updates are projected to the tangent space, resulting in the ODE given by
\begin{equation}
\gdd = -\|\gd\|^2 \left( I - \hat{\gamma} \hat{\gamma}\transpose \right) \left( I - \hat{\gd} \hat{\gd}\transpose \right) \nabla \log p(\gamma),
\label{eq:ivp_acc}
\end{equation}
which can be solved by applying the Runge--Kutta (RK4) method to the first-order system of equations.

\section{Experiments}
\label{sec:experiments}

In this section, we first analyze whether short video sequences are geodesics in diffusion latent space.
Second, we evaluate applications of the theory, focusing on the image interpolation task framed as solving a boundary value problem.
For all experiments, we use a pre-trained Stable Diffusion v2.1-base model \cite{rombach2022high}.

\subsection{Geodesic analysis of videos}
\label{sec:analysis}

The objective of this experiment is to assess how close video clips are to geodesics.

\paragraph{Dataset.}
We use the CLEVR framework \cite{johnson2017clevr} to render 40 synthetic videos with Blender. 
Each 100-frame sequence contains one of the three CLEVR objects (cube, sphere, cylinder) and has a single varying attribute, including camera rotation and translation, light source location, and the size, color and motion of the object.

\paragraph{Baselines.}
The path $\gamma_v$ corresponding to the original video clip is compared to
the path after geodesic optimization $\gamma_o$, 
the path after a sinuoidal perturbation $\gamma_\delta = \gamma_v + \delta\sin{(\pi\gamma)}$, and
the path after smoothing $\gamma_s$, by fitting a smoothed cubic spline \cite{de1978practical} with smoothing factor 0.9999.

\begin{figure}[!t]\centering
    \begin{subfigure}[]{\linewidth}\centering\small
        \input{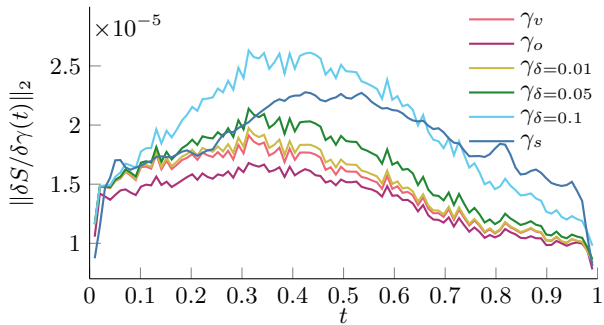}
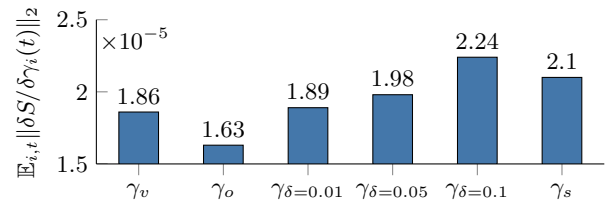%
        \caption{Geodesic gradient norm along the path, for the translating red cube.
        }
        \label{fig:clevr_norms}
    \end{subfigure}\vfill 
    \begin{subfigure}[]{\linewidth}\centering\small
        \begin{tikzpicture}
\begin{axis}[
ybar,                                  
bar width=15pt,                        
width=\textwidth,                      
height=3.5cm,                            
ymin=1.5,                                
ymax=2.5,                               
symbolic x coords={
$\gamma_v$, 
$\gamma_o$, 
$\gamma_{\delta=0.01}$, 
$\gamma_{\delta=0.05}$, 
$\gamma_{\delta=0.1}$, 
$\gamma_s$
},     
xtick=data,                            
ytick={1.5, 2, 2.5},
nodes near coords,                     
ylabel={$\mathbb{E}_{i,t}\|\delta S / \delta \gamma_i (t) \|_2$},                       
axis x line*=bottom,
axis y line*=left,
]

\node[anchor=north west] at (rel axis cs:-0.01,1) {\small $\times 10^{-5}$};

\definecolor{purple}{RGB}{170, 51, 119}
\definecolor{pink}{RGB}{238, 102, 119}
\definecolor{yellow}{RGB}{204, 187, 68}
\definecolor{green}{RGB}{34, 136, 51}
\definecolor{cyan}{RGB}{102, 204, 238}
\definecolor{blue}{RGB}{68, 119, 170}


\addplot[fill=blue] coordinates {
($\gamma_v$,1.86)
($\gamma_o$,1.63) 
($\gamma_{\delta=0.01}$,1.89)
($\gamma_{\delta=0.05}$,1.98) 
($\gamma_{\delta=0.1}$,2.24) 
($\gamma_s$, 2.10)};
\end{axis}
\end{tikzpicture}%
        \caption{Geodesic gradient norm, averaged across the path and dataset.}
        \label{fig:clevr_avg_norms}
    \end{subfigure}%
    \caption{
        Analysis of simple videos generated using CLEVR \cite{johnson2017clevr}.
        If a video path was a geodesic in the diffusion latent space, we would expect (i) the norm of the geodesic gradient along the path $\gamma_v$ to be near zero, (ii) the norm to be close to that of the optimized path $\gamma_o$, and (iii) the norm of the perturbed $\gamma_\delta$ or smoothed $\gamma_s$ paths to be larger.
        From the evidence, we conclude that many of the videos are approximately geodesic.
        }
    \label{fig:clevr}
\end{figure}

\paragraph{Results.}
In \cref{fig:clevr_norms}, we report the $l_2$ norm of the functional derivative in \cref{eq:derivative} as it varies along the path, for a single video clip (translating red cube).
For a geodesic, this is zero everywhere along the path.
In \cref{fig:clevr_avg_norms}, we report the average of this norm across the path and the dataset.
If a video path was a geodesic in the diffusion latent space, we would expect (i) the norm along the path $\gamma_v$ to be near zero, (ii) the norm to be close to that of the optimized path $\gamma_o$, and (iii) the norm of the perturbed $\gamma_\delta$ or smoothed $\gamma_s$ paths to be larger.
We observe these trends in the results, and conclude that many of the videos are approximately geodesic.

\subsection{Applications}
\label{sec:exp-app}
Here, we assess the performance of the geodesic solvers for the image interpolation and extrapolation tasks.

\paragraph{Datasets.}
\label{exp:datasets}
We compile a union of datasets from prior works \cite{zhang2024diffmorpher, yang2024impus, wang2023interpolating}. 
This includes MorphBench \cite{zhang2024diffmorpher}, which contains 90 image pairs of object animations and object metamorphoses;
Animals and Humans \cite{yang2024impus}, which contains 50 animal image pairs from AFHQ \cite{choi2020stargan} with an LPIPS below 0.7 and 50 human face image pairs from CelebA-HQ \cite{karras2017progressive} with an LPIPS below 0.6; 
and Web, which contains 20 image pairs sourced from publicly accessible websites, some of which have been used in other related studies \cite{wang2023interpolating}.

\paragraph{Metrics.}
Quantitatively evaluating image interpolation and extrapolation is extremely challenging and subjective.
Following previous work \cite{zhang2024diffmorpher, yang2024impus},
we report
(1)~the Fr\'{e}chet inception distance (FID) \cite{heusel2017gans} between the set of input images and the set of generated images to measure how close the distributions are;
(2)~the perceptual path length (PPL) \cite{karras2019style} as the sum of LPIPS between adjacent images to assess the directness of the generated image sequence;
(3)~perceptual distance variance (PDV) as the standard deviation of LPIPS between consecutive images to assess the consistency of transition rates across the sequence; and
(4)~the TOPIQ \cite{chen2024topiq} score, an image quality metric that evaluates the perceptual quality of each generated image in alignment with human perception. 
To focus on the quality of interior frames,
we compute a weighted TOPIQ score, detailed in \cref{sec:app_eva}.
We report all metrics on 17-frame sequences by sampling 15 intermediate images between the input pairs.

\begin{figure}[!t]\centering
    \raisebox{0.75em}{\rotatebox{90}{\scriptsize NoiseD.}}\hfill
    \includegraphics[width=0.97\linewidth]{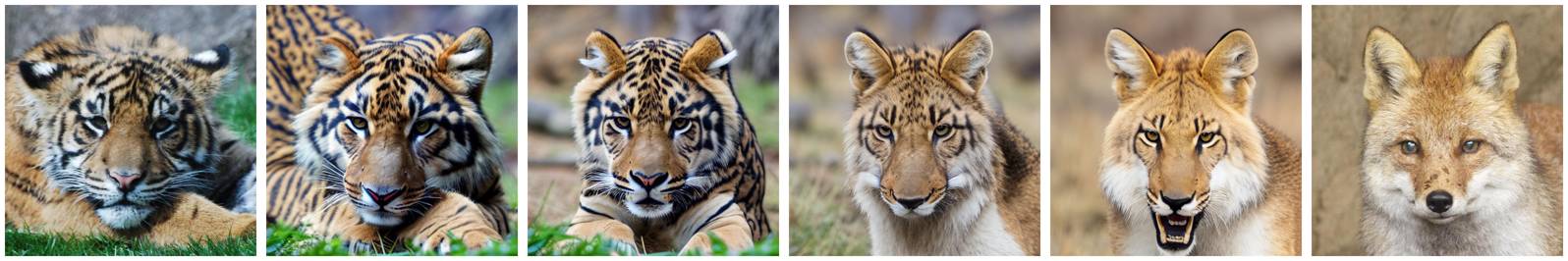}\vfill
    \raisebox{1.25em}{\rotatebox{90}{\scriptsize AID}}\hfill
    \includegraphics[width=0.97\linewidth]{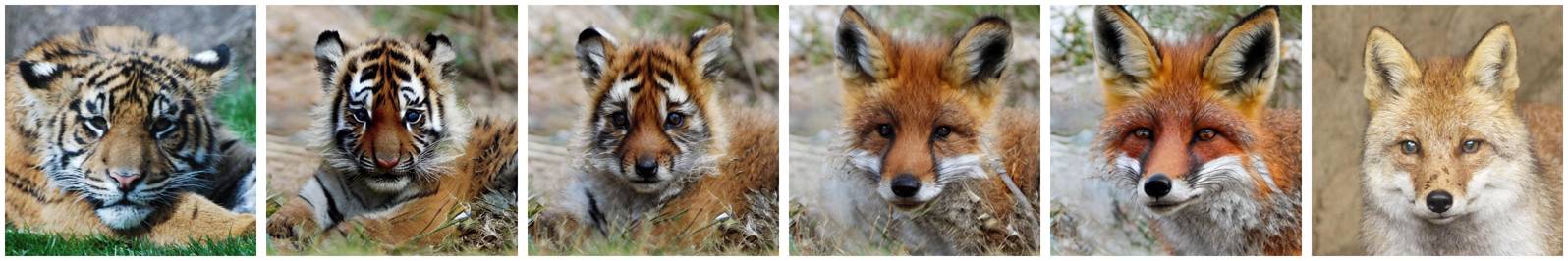}\vfill
    \raisebox{1em}{\rotatebox{90}{\scriptsize DiffM.}}\hfill
    \includegraphics[width=0.97\linewidth]{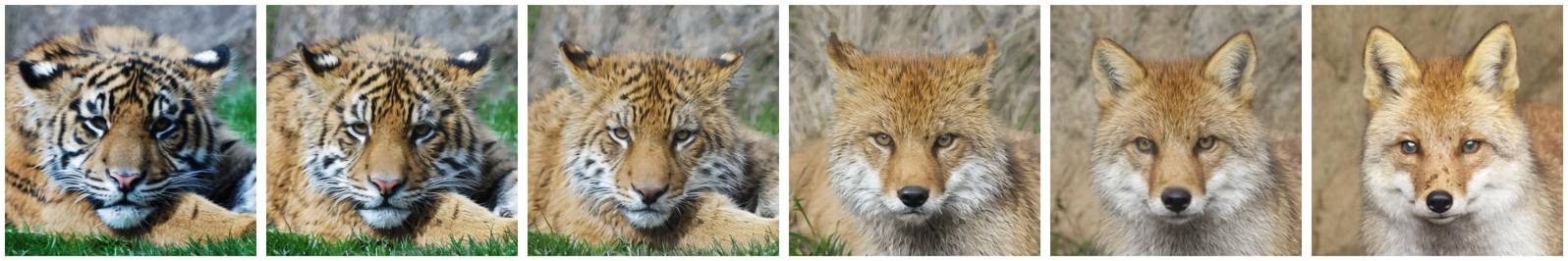}\vfill
    \raisebox{1em}{\rotatebox{90}{\scriptsize IMPUS}}\hfill
    \includegraphics[width=0.97\linewidth]{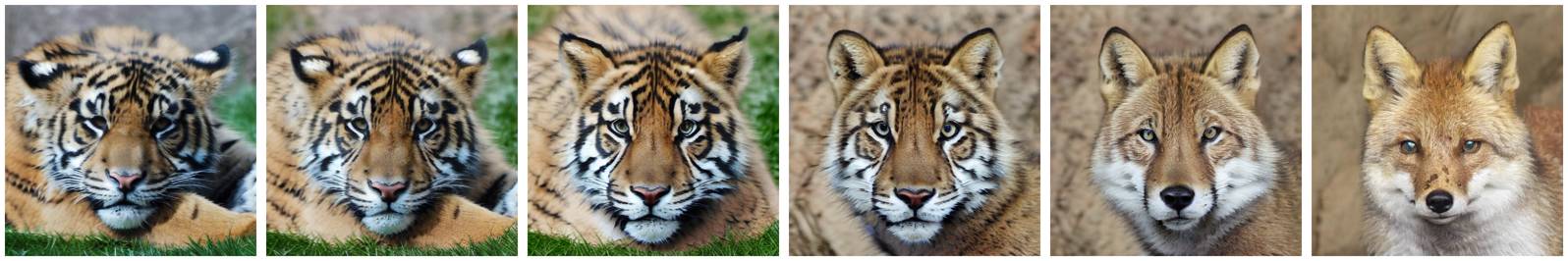}\vfill
    \raisebox{0.75em}{\rotatebox{90}{\scriptsize SmoothD.}}\hfill
    \includegraphics[width=0.97\linewidth]{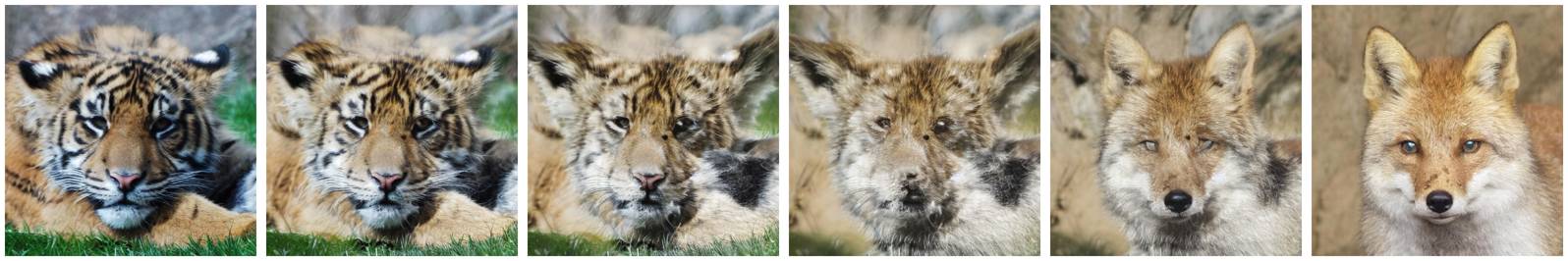}\vfill
    \raisebox{0.1em}[0pt]{\rotatebox{90}{\scriptsize Ours w/o opt.}}\hfill
    \includegraphics[width=0.97\linewidth]{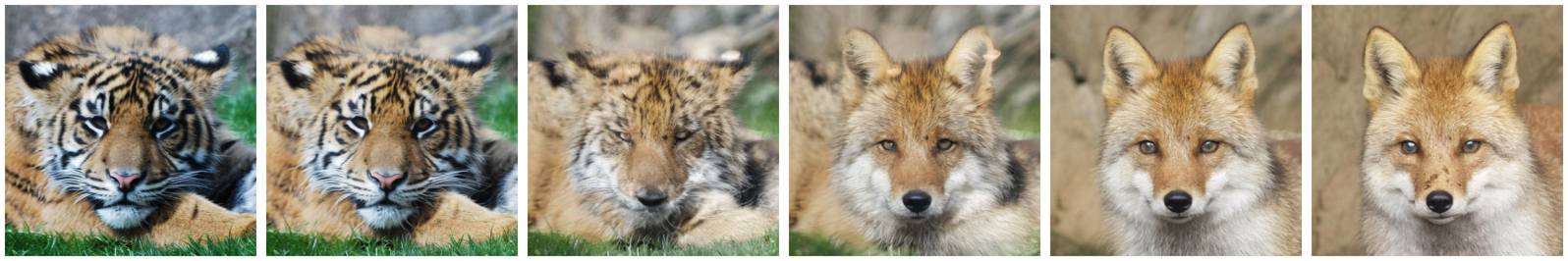}\vfill
    \raisebox{1.4em}{\rotatebox{90}{\scriptsize Ours}}\hfill
    \includegraphics[width=0.97\linewidth]{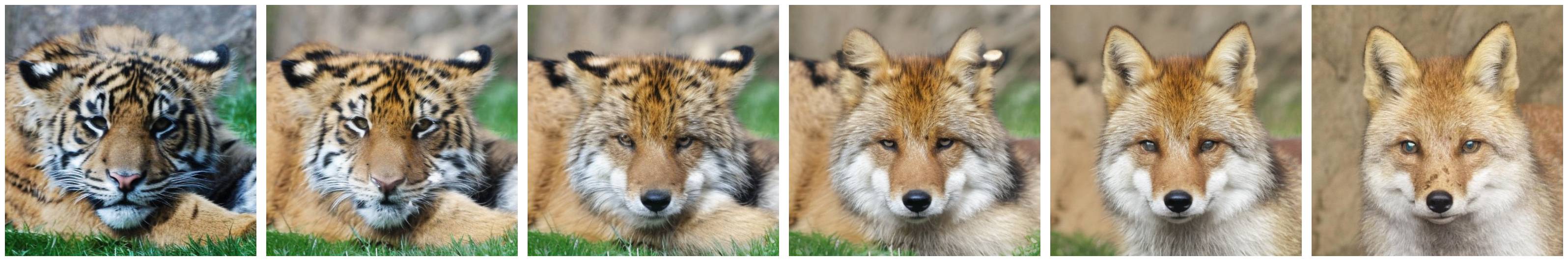}%
    \caption{
    Qualitative comparison of image interpolation results. 
    }
    \label{fig:qual_comparison}
\end{figure}

\paragraph{Implementation details.}
We use BLIP \cite{li2022blip} to generate short text prompts $p$ for each input image.  
For text inversion, we finetune the CLIP \cite{radford2021learning} encoded text embedding $z$, using 500 steps with a learning rate of 0.005 and the AdamW optimizer. 
For BVP optimization, we perform 400 steps of gradient descent with an initial learning rate of 0.1, following a linear learning rate schedule.
The hyperparameters are set as $\tau=600$, $\Delta \tau=100$, $\beta=0.002$, and $\sigma=1$. 
As detailed in \cref{sec:method_alg}, we adopt a bisection strategy to add additional sample points every 100 steps.  
For IVP optimization, we set the iteration number to 200.
To obtain a diverse set of initial velocities, we apply text inversion to the target text embedding using several generated images from random initial noise $\mathcal{N}(0, I)$, conditioned on the target prompt \cite{ruiz2023dreambooth}.  
All methods use the same pre-trained diffusion model (Stable Diffusion v2.1-base) and text prompts, and default settings otherwise. 
Further details are in \cref{sec:app_hyperparameters}.

\begin{table*}[!t]
    \centering
    \small
    \setlength{\tabcolsep}{1pt} 
    \caption{
    Image interpolation results.
    We report the performance on three datasets: MorphBench (MB), Animals and Humans (AH), and Web data (Web). 
    TF denotes ``training-free'', which also prohibits fine-tuning; best results are in bold; second best ones are underlined.
    ``Ours'' represents the optimized geodesic path and
    ``Ours w/o opt.'' represents the unoptimized path as the great circle initialization.  
    }
    \label{tab-main}
    \renewcommand{\arraystretch}{1.1} 
    \vspace{-6pt}
    \begin{tabularx}{ \linewidth}{@{}ll CCC CCC CCC CCC@{}}
        \toprule
        &
        & \multicolumn{3}{c}{ FID $\downarrow$} 
        & \multicolumn{3}{c}{ PPL$\downarrow$}
        & \multicolumn{3}{c}{ PDV$\downarrow$} 
        & \multicolumn{3}{c}{ TOPIQ $\uparrow$}
        \\
        \cmidrule(lr){3-5} \cmidrule(lr){6-8} \cmidrule(lr){9-11} \cmidrule(lr){12-14}
        Method  & TF
        & AH & MB & Web 
        & AH & MB & Web    
        & AH & MB & Web
        & AH & MB & Web\\
        \midrule
        NoiseDiffusion \cite{zheng2024noisediffusion} 
        & \checkmark
        & 71.85 & 100.87 & 201.08
        & 2.507 & 2.718 & 3.738
        & 0.117 & 0.111 & 0.099 
        & \textbf{0.700} & \underline{0.666} & \underline{0.650}
        \\
        AID \cite{he2024aid}
        & \checkmark
        & 86.30 & 124.11 & 242.34
        & 2.709 & 2.648 & 3.813  
        & 0.181 & 0.188 & 0.208
        & \underline{0.699} & \textbf{0.675} & \textbf{0.665}
        \\
        IMPUS \cite{yang2024impus} 
        & \xmark
        & \underline{25.75} & \textbf{36.33} & \textbf{89.73}
        & 1.861 & 1.718 & 2.554  
        & 0.065 & 0.066 & 0.115
        & 0.622 & 0.587 & 0.539
        \\
        DiffMorpher \cite{zhang2024diffmorpher} 
        & \xmark
        & 32.89 & 42.39 & 160.69
        & 1.195 & 1.011 & 1.934  
        & \underline{0.018} & \textbf{0.016} & \textbf{0.024}
        & 0.686 & 0.651 & 0.592
        \\
        SmoothDiffusion \cite{guo2024smooth}
        & \xmark
        & 30.80 & 52.19 & 135.78
        & \underline{0.903} & \underline{0.879} & \textbf{1.371}
        & 0.027 & 0.033 & 0.045
        & 0.571 & 0.515 & 0.389\\
        \midrule
        Ours w/o opt.
        & \checkmark
        & \textbf{24.71} & \underline{37.85} & \underline{112.81} 
        & \textbf{0.874} & \textbf{0.841} & \underline{1.473} 
        & 0.032 & 0.035 & 0.053
        & 0.584 & 0.546 & 0.466 \\
        Ours
        & \checkmark
        & 33.87 & 46.51 & 134.68
        & 0.960 & 0.921 & 1.565 
        & \textbf{0.016} & \underline{0.022} & \underline{0.026} 
        & 0.607 & 0.559 & 0.479 \\
        \bottomrule  
    \end{tabularx}
\end{table*}

\begin{figure*}[!t]\centering
     \begin{subfigure}[]{\linewidth}\centering 
        \includegraphics[width=\linewidth]{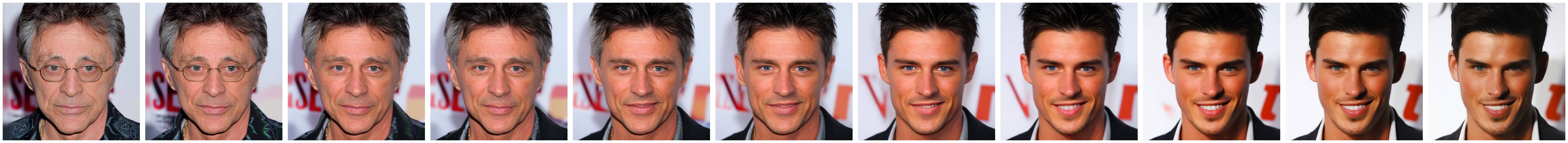}
    \end{subfigure}\vfill
      \begin{subfigure}[]{\linewidth}\centering 
        \includegraphics[width=\linewidth]{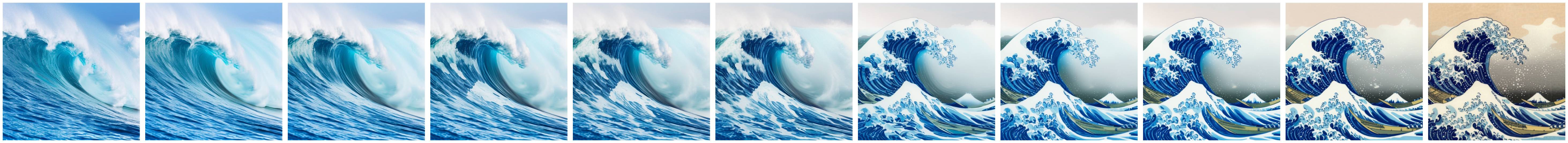}
    \end{subfigure}
    \begin{subfigure}[]{\linewidth}\centering 
        \includegraphics[width=\linewidth]{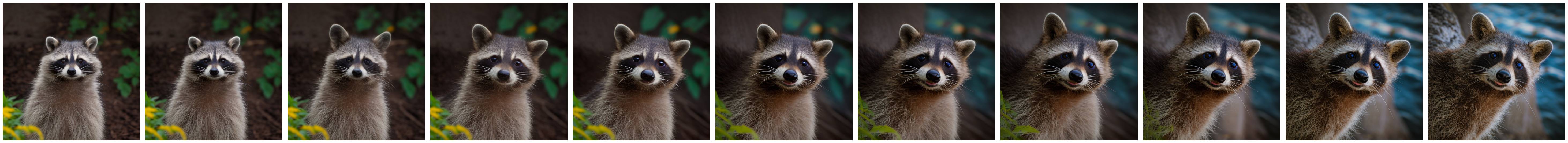}
    \end{subfigure}\vfill
    \begin{subfigure}[]{\linewidth}\centering 
        \includegraphics[width=\linewidth]{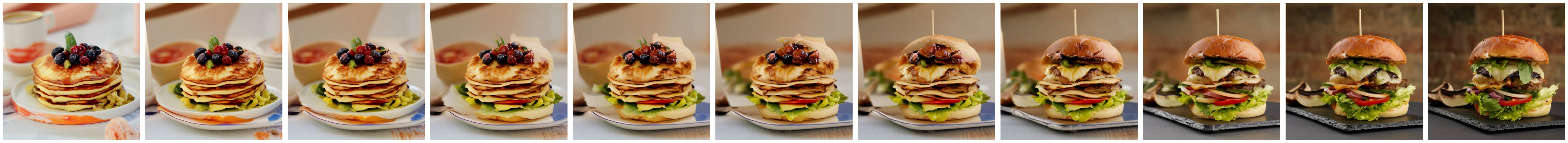}
    \end{subfigure}\vfill
    \begin{subfigure}[]{\linewidth}\centering 
        \includegraphics[width=\linewidth]{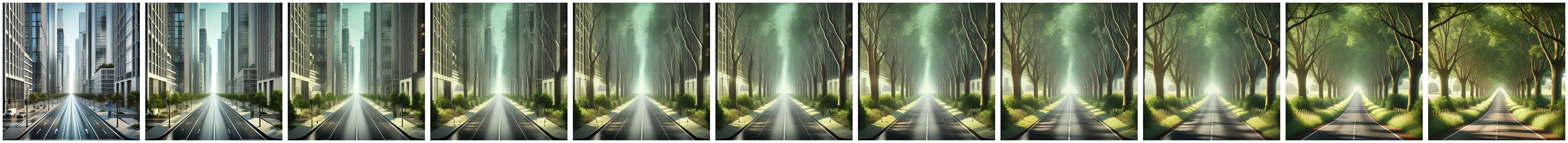}
    \end{subfigure}\vfill
    \caption{
        Qualitative image interpolation results using our geodesic BVP solver. \label{fig:qualitative}
	}    
\end{figure*}

\begin{figure}[!t]\centering
    \begin{subfigure}[]{\linewidth}\centering 
        \includegraphics[width=\linewidth]{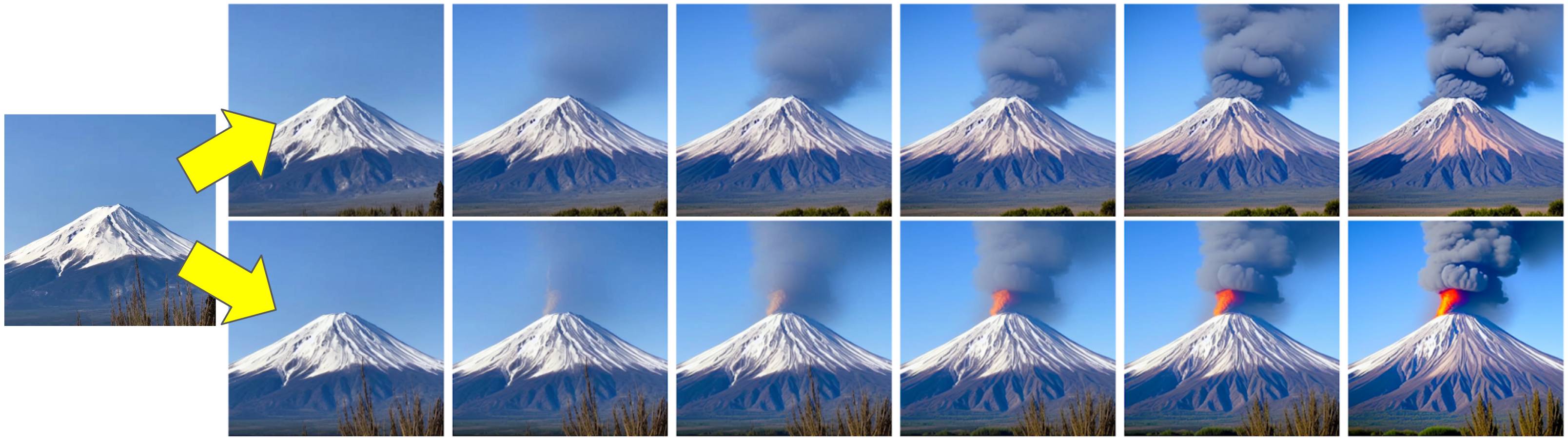}
        \caption{Target prompt: ``Volcano eruption''.}
        \label{fig:volcano}
    \end{subfigure}\vfill
    \begin{subfigure}[]{\linewidth}\centering 
        \includegraphics[width=\linewidth]{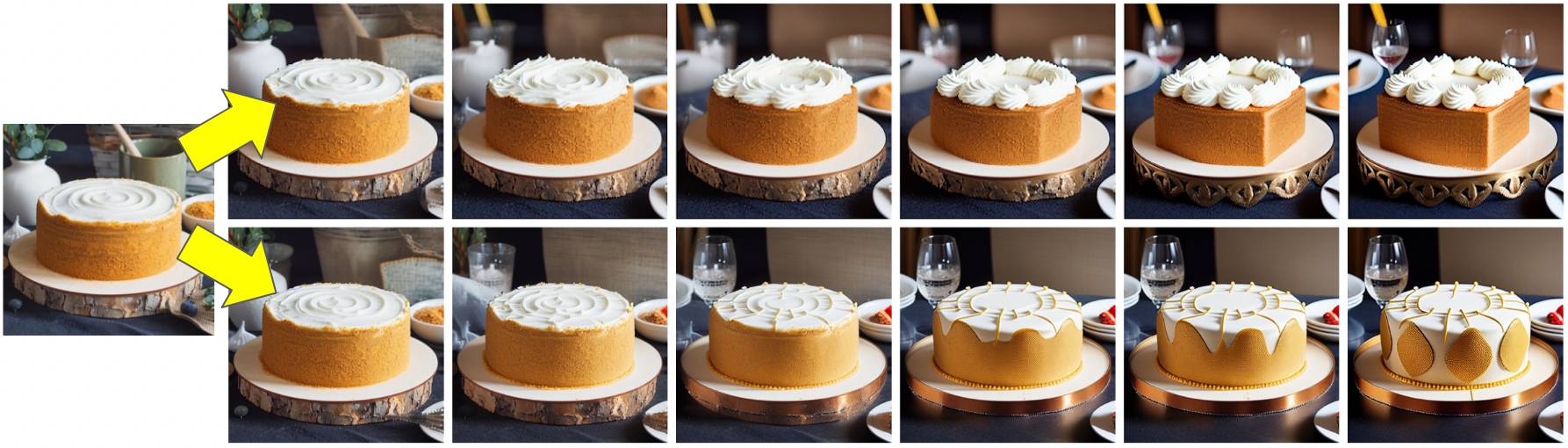}
        \caption{Target prompt: ``An expensive cake worth \$1,000, adorned with fancy decorations and looking incredibly delicious''.}
        \label{fig:cake}
    \end{subfigure}\\
    \begin{subfigure}[]{\linewidth}\centering 
        \includegraphics[width=\linewidth]{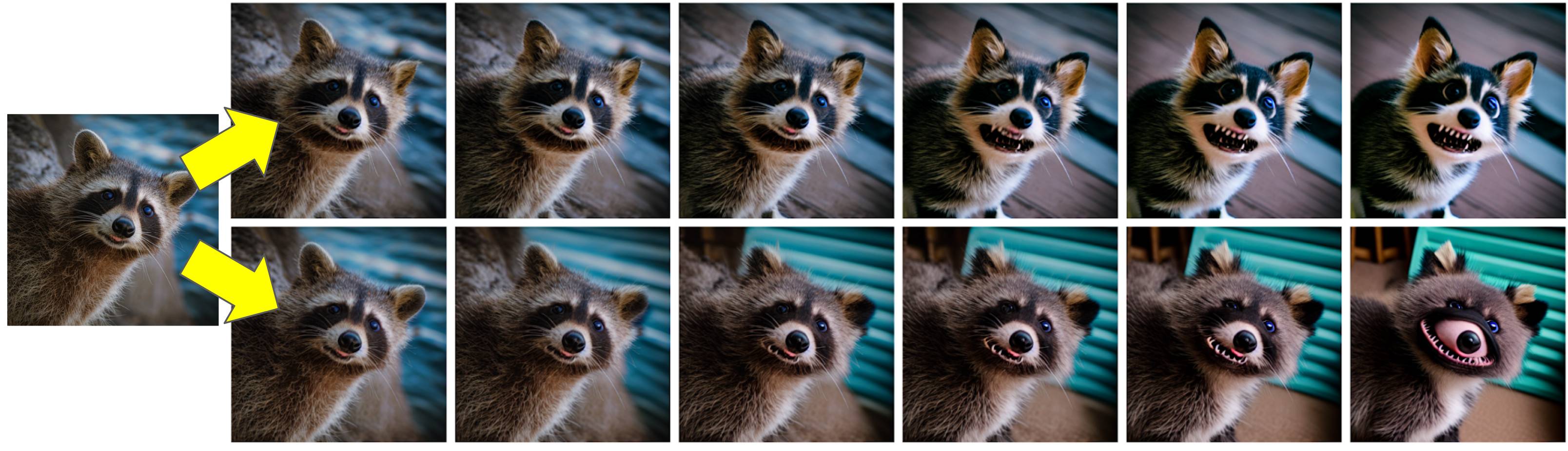}
        \caption{Target prompt: ``A real photo of a cute but scary monstrous creature with sharp teeth and bulging eyes''.}
        \label{fig:raccoon}
    \end{subfigure}%
    \caption{
        Qualitative image extrapolation results using our geodesic IVP solver. For each image, we plot two extrapolated paths, each with different initial velocities but the same prompt.
	}
    \label{fig:ivp}
\end{figure}

\paragraph{Results.}
We compare our method for the image interpolation task with several state-of-the-art diffusion-based methods, including
NoiseDiffusion \cite{zheng2024noisediffusion},
DiffMorpher \cite{zhang2024diffmorpher}, 
IMPUS \cite{yang2024impus},
AID \cite{he2024aid}, 
SmoothDiffusion \cite{guo2024smooth}
and report the result of a baseline (`Ours w/o opt.'), which returns the initial path before geodesic optimization.
This corresponds to the great circle trajectory, the most direct path between the endpoints on the sphere.
The quantitative and qualitative results are presented in \cref{tab-main,fig:qual_comparison,fig:qualitative}.
We observe that methods involving finetuning the diffusion model on input images, like DiffMorpher and IMPUS, tend to score better with respect to most of the metrics. 
While AID and NoiseDiffusion generate high-quality images, they have a weaker connection to the input images.
SmoothDiffusion \cite{guo2024smooth}, trained on the large LAION dataset, performs well with respect to perceptual path length, but has high variance and weaker image quality scores, especially on the partially OOD Web dataset.
In contrast, our method has high directness (low PPL), high fidelity to the input distribution (low FID), and very high perceptual smoothness (low PDV), but has slightly lower image quality (low TOPIQ).
Overall, our approach is on-par with the best methods without requiring any training.
Finally, we show qualitative results for the image extrapolation task in \cref{fig:ivp}, where we visualize two trajectories for each prompt.

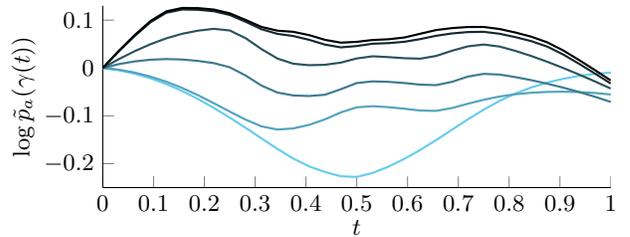
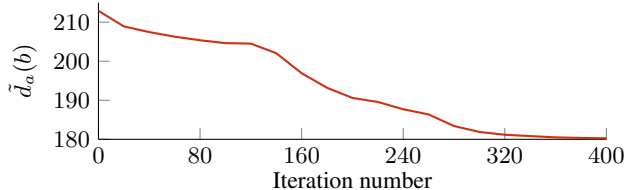
\begin{figure}[!t]\centering
    \begin{subfigure}[]{\linewidth}\centering\small
\begin{tikzpicture}
\begin{axis}[
width=\textwidth,
height=4.0cm,
xmin=0,
xmax=1,
xtick={0, 0.1, 0.2, 0.3, 0.4, 0.5,0.6, 0.7, 0.8,0.9, 1},
xlabel={$t$},
xlabel shift=-3pt,
ymin=-0.25,
ymax=0.13,
ytick={-0.2, -0.1, 0, 0.1},
ylabel={$\log \tilde{p}_a(\gamma(t))$},
ylabel shift=-3pt,
axis x line*=bottom,
axis y line*=left,
]


\definecolor{c1}{RGB}{102, 204, 238}  
\definecolor{c2}{RGB}{82, 163, 190}   
\definecolor{c3}{RGB}{61, 122, 143}   
\definecolor{c4}{RGB}{41, 82, 95}    
\definecolor{c5}{RGB}{20, 41, 48}    
\definecolor{c6}{RGB}{0, 0, 0}  

\addplot [c1,thick] table [row sep=newline]{
0.0 0.0
0.03125 -0.0046661384
0.0625 -0.01179354
0.09375 -0.021397814
0.125 -0.03347735
0.15625 -0.04800396
0.1875 -0.06482648
0.21875 -0.08356369
0.25 -0.10377134
0.28125 -0.12510985
0.3125 -0.14717054
0.34375 -0.16910698
0.375 -0.18927811
0.40625 -0.20563042
0.4375 -0.21748644
0.46875 -0.22637719
0.5 -0.22775474
0.53125 -0.21943069
0.5625 -0.20621158
0.59375 -0.18987668
0.625 -0.17126817
0.65625 -0.15075237
0.6875 -0.12904437
0.71875 -0.10743287
0.75 -0.0869971
0.78125 -0.0687184
0.8125 -0.053247426
0.84375 -0.04061051
0.875 -0.030532688
0.90625 -0.022634106
0.9375 -0.016595354
0.96875 -0.01222185
1.0 -0.009432956
};

\addplot [c2,thick] table [row sep=newline]{
0.0 0.0
0.03125 -0.0034870622
0.0625 -0.009468305
0.09375 -0.017877977
0.125 -0.028634764
0.15625 -0.041708518
0.1875 -0.056906246
0.21875 -0.07366915
0.25 -0.09107879
0.28125 -0.107935406
0.3125 -0.12201573
0.34375 -0.1286885
0.375 -0.12632953
0.40625 -0.11889795
0.4375 -0.107370295
0.46875 -0.09279423
0.5 -0.082268424
0.53125 -0.079878174
0.5625 -0.081871256
0.59375 -0.08511151
0.625 -0.088978395
0.65625 -0.089633375
0.6875 -0.08375798
0.71875 -0.07467035
0.75 -0.06639554
0.78125 -0.05966192
0.8125 -0.054824766
0.84375 -0.05172555
0.875 -0.050030075
0.90625 -0.049563635
0.9375 -0.050298154
0.96875 -0.052233323
1.0 -0.055372227
};

\addplot [c3,thick] table [row sep=newline]{
0.0 0.0
0.03125 0.007995306
0.0625 0.013724325
0.09375 0.017287303
0.125 0.018659404
0.15625 0.017780198
0.1875 0.014784209
0.21875 0.010202469
0.25 0.001459456
0.28125 -0.015735332
0.3125 -0.036771372
0.34375 -0.051322464
0.375 -0.057373736
0.40625 -0.05859365
0.4375 -0.055919506
0.46875 -0.044371627
0.5 -0.031493858
0.53125 -0.027949564
0.5625 -0.028954197
0.59375 -0.03139479
0.625 -0.034814198
0.65625 -0.035820175
0.6875 -0.028931158
0.71875 -0.01799775
0.75 -0.012248011
0.78125 -0.013728667
0.8125 -0.019066751
0.84375 -0.025466442
0.875 -0.032807056
0.90625 -0.04102533
0.9375 -0.050121814
0.96875 -0.06009353
1.0 -0.070931405
};

\addplot [c4,thick] table [row sep=newline]{
0.0 0.0
0.03125 0.018350068
0.0625 0.034938525
0.09375 0.049638364
0.125 0.061996803
0.15625 0.071541175
0.1875 0.07808343
0.21875 0.08190921
0.25 0.07845174
0.28125 0.062297903
0.3125 0.03944454
0.34375 0.019848865
0.375 0.008927893
0.40625 0.0058400016
0.4375 0.006684512
0.46875 0.011303332
0.5 0.019951466
0.53125 0.024486363
0.5625 0.02273027
0.59375 0.020459805
0.625 0.019465968
0.65625 0.02516579
0.6875 0.03606401
0.71875 0.04525451
0.75 0.049023677
0.78125 0.04482362
0.8125 0.035694443
0.84375 0.024986774
0.875 0.013021318
0.90625 8.212915e-05
0.9375 -0.01363063
0.96875 -0.027995914
1.0 -0.042977978
};

\addplot [c5,thick] table [row sep=newline]{
0.0 0.0
0.03125 0.034318816
0.0625 0.06644766
0.09375 0.09546782
0.125 0.11541455
0.15625 0.12223995
0.1875 0.12165523
0.21875 0.11858707
0.25 0.11080194
0.28125 0.09666216
0.3125 0.0810695
0.34375 0.071398735
0.375 0.066828854
0.40625 0.0588934
0.4375 0.048896424
0.46875 0.043109663
0.5 0.045877106
0.53125 0.050719
0.5625 0.052002266
0.59375 0.05478295
0.625 0.06042123
0.65625 0.067348294
0.6875 0.07224167
0.71875 0.07513283
0.75 0.075586095
0.78125 0.07158673
0.8125 0.06448021
0.84375 0.05611432
0.875 0.044719964
0.90625 0.028336938
0.9375 0.008862473
0.96875 -0.011457784
1.0 -0.032456093
};

\addplot [c6, thick] table [row sep=newline]{
0.0 0.0
0.03125 0.035477515
0.0625 0.06850615
0.09375 0.098100714
0.125 0.11838609
0.15625 0.12551458
0.1875 0.12515543
0.21875 0.12211118
0.25 0.11431317
0.28125 0.10051326
0.3125 0.08584821
0.34375 0.07807332
0.375 0.07564368
0.40625 0.068735436
0.4375 0.058882143
0.46875 0.052970283
0.5 0.05465857
0.53125 0.058562312
0.5625 0.060078286
0.59375 0.06377933
0.625 0.07157517
0.65625 0.079363585
0.6875 0.08338618
0.71875 0.085734785
0.75 0.085794345
0.78125 0.08151346
0.8125 0.074137226
0.84375 0.065444976
0.875 0.053658392
0.90625 0.03680099
0.9375 0.016738128
0.96875 -0.0042640744
1.0 -0.02601496

};

\end{axis}
\end{tikzpicture}%
     \caption{Relative log-probability along the path as optimization progresses, where iteration 0 is the lightest curve and iteration 400 is the darkest.}
    \end{subfigure}\vfill
    \begin{subfigure}[]{\linewidth}\centering\small
\begin{tikzpicture}
\begin{axis}[
width=\textwidth,
height=3.4cm,
xmin=0,
xmax=400,
xtick={0, 80, 160, 240, 320, 400},
xlabel={Iteration number},
xlabel shift=-3pt,
ymin=180,
ymax=215,
ytick={180, 190, 200, 210},
ylabel={$\tilde{d}_a(b)$},
axis x line*=bottom,
axis y line*=left,
]

\definecolor{c}{RGB}{204,51,17}

\addplot [c, thick] table [row sep=newline]{
0 212.89703
20 208.9169
40 207.45642
60 206.28667
80 205.36386
100 204.62416
120 204.50775
140 202.05498
160 196.92186
180 193.1952
200 190.59125
220 189.56627
240 187.69463
260 186.36917
280 183.39618
300 181.88391
320 181.1651
340 180.81494
360 180.49706
380 180.33922
400 180.25125
};




\end{axis}
\end{tikzpicture}%
     \caption{Path length with respect to optimization step. 
     }
    \end{subfigure} 
    \caption{
        Example of the evolution of the probability density along the path and the path length during BVP optimization.
	}
    \label{fig:opt}
\end{figure}

\begin{table}[!t]
    \centering
    \small
    \setlength{\tabcolsep}{1pt} 
    \renewcommand{\arraystretch}{1.1} 
    \caption{
        We ablate the impact of different text conditioning strategies for BVP solver, including text inversion, positive ($\oplus$) prompts, and negative ($\ominus$) prompts, on the validation dataset.}
    \label{tab-morph_ablation}
    \vspace{-6pt}
    \begin{tabularx}{\linewidth}{@{}cccCCCC@{} } 
    \toprule
    Text inv.
    & $\oplus$ prompt
    & $\ominus$ prompt
    & FID$\downarrow$
    & PPL$\downarrow$
    & PDV$\downarrow$
    & TOPIQ$\uparrow$ \\
    \midrule
     & $\checkmark$ &  & 70.56 & 1.06 & 0.017 & 0.518\\
     &  & $\checkmark$  & 69.06 & 1.046 & 0.017 & 0.529  \\
    $\checkmark$& $\checkmark$ &  & 65.05 & 1.045 & 0.018 & 0.562\\
    & $\checkmark$ & $\checkmark$ & 64.72 & 1.037 & 0.019 & \textbf{0.564}\\
    $\checkmark$ & $\checkmark$ & $\checkmark$ & \textbf{63.28} & \textbf{1.000} & \textbf{0.017} & 0.553  \\
    \bottomrule    
    \end{tabularx}
\end{table}

\paragraph{Analysis.}
We constructed a validation dataset for ablation and analysis by randomly selecting 25 image pairs from the three datasets (\cref{exp:datasets}). 
In \cref{fig:opt}, we show that optimization correctly increases the log probability of samples along the path and decreases the path length.
This indicates that the curve smoothly approaches a geodesic during optimization.
We also present an ablation study in \cref{tab-morph_ablation}, where we show how the text conditioning formulation (text inversion, positive prompt, and negative prompt) contributes to the performance of the method. 
In \cref{sec:app_cond}, we compare the time-linear conditioning signal with the constant one.
In \cref{sec:app_anahyper}, we analyze the trade-off between the different choices of hyperparameter settings.

\subsection{Limitations}

This work has several limitations.
First, the metrics used for evaluating image interpolation struggle to capture interpolation quality (smoothness, directness, and realism).
In particular, FID, which measures fidelity to the input distribution, is unreliable when applied to small image sets like these, and
the other metrics only capture individual aspects of interpolation quality. 
Second, the method performs well for image morphing (local changes) but struggles with large camera motions or domain gaps, as illustrated in \cref{fig:interp_failure_1,fig:interp_failure_2}. 
We hypothesize that by initializing with a great circle, which also performs poorly in such cases, our optimizer gets stuck in nearby local optima rather than finding more optimal solutions.
A global search strategy is indicated.
Third, we approximate the log-probability gradient with a score distillation gradient, which is not necessarily well-aligned.
Fourth, the path representation is inelegant: a spherical piecewise linear function (great circle arcs) for which the velocities and accelerations are approximated by fitting a cubic spline.
Finally, the image extrapolation performance is unreliable, as determining a good initial velocity is challenging;
a more robust approach is called for.

\section{Conclusion}
\label{sec:conclusion}
In this paper, we have presented the theory required for computing probability density geodesics in diffusion latent space and algorithms for solving the associated initial and boundary value problems.
We also described how to compute several useful quantities for analysis: the relative probability density along the path, the geodesic distance between two points, and the geodesic gradient norm.
Finally, we presented applications to image interpolation and extrapolation and evaluated the performance of these training-free approaches.
We show that they perform comparably or better than existing state-of-the-art.
We expect these techniques to be useful for tasks involving generative modeling, as well as for studying the distribution of image space.

\newpage
\section*{Acknowledgments}
We thank all reviewers and ACs for their constructive comments.
This research was, in part, funded by the U.S.\ Government---DARPA TIAMAT HR00112490421. The views and conclusions contained in this document are those of the authors and should not be interpreted as representing the official policies, either expressed or implied, of the U.S.\ Government.

{
    \small
    \bibliographystyle{ieeenat_fullname}
    \bibliography{main}
}

\clearpage
\setcounter{page}{1}
\maketitlesupplementary
\appendix

In this appendix, we provide the full derivations for the mathematical results presented in the main paper and additional (especially qualitative) experimental results.

\section{Derivations}
\label{sec:app_derivations}

In this section, we provide the derivations deferred from the main paper.
We formulate the problem in more general terms---as a weighted path length---before returning to the specifics used in the main paper.

\paragraph{Equation for path length.}
Let $\gamma : [a, b] \rightarrow \R^n$ be a path such that $\gamma(a) = \vx_a$ and $\gamma(b) = \vx_b$, and $S : \{\gamma_i\} \rightarrow \R$ be the action functional on the set of such paths, defined by
\begin{align}
S[\gamma] &= \int_a^b L(t, \gamma(t), \gd(t)) \dt,
\end{align}
where $L$ is the Lagrangian given by
\begin{align}
L(t, \gamma(t), \gd(t)) &= \| \gd(t) \| w(\gamma(t)).\\
L(t, \gamma(t), \gd(t)) &= \sqrt{ \langle \gd(t), \gd(t) \rangle}_{K(\gamma(t))} \\
K(\gamma(t)) &= w(\gamma(t))^2 I.
\end{align}
Then $S[\gamma]$ is the weighted path length for path $\gamma$.

\paragraph{Euler--Lagrange equations.}
Given this definition, a path $\gamma$ is a stationary point of $S$ iff it satisfies the Euler--Lagrange equations, \viz,
\begin{align}
\pp{}{\gamma} L(t, \gamma(t), \gd(t)) - \ddt{}\pp{}{\gd} L(t, \gamma(t), \gd(t)) &= 0,
\end{align}
where $\pp{L}{\gamma}$ stacks the partial derivatives \wrt the components of $\gamma$ and $\pp{L}{\gd}$ stacks the partial derivatives \wrt the components of $\gd$.
We obtain
\begin{align}
0 &= \pp{}{\gamma} L(t, \gamma(t), \gd(t)) - \ddt{}\pp{}{\gd} L(t, \gamma(t), \gd(t))\\
&= \pp{}{\gamma} \left(\| \gd(t) \| w(\gamma(t)) \right) - \ddt{}\pp{}{\gd} \left( \| \gd(t) \| w(\gamma(t))\right)\\
&= \| \gd \| \dd{w}{\gamma} - \ddt{}\left( \frac{\gd}{\| \gd \|} w(\gamma(t)) \right)\\
&= \| \gd \| \dd{w}{\gamma} - \ddt{}\left( \frac{\gd}{\| \gd \|} \right) w(\gamma)  - \frac{\gd}{\| \gd \|} \ddt{}\left(w(\gamma(t)) \right)\\
&= \| \gd \| \dd{w}{\gamma} - \frac{1}{\| \gd \|} \left( I - \frac{\gd \gd\transpose}{\| \gd \|^2} \right) \gdd w  - \frac{\gd}{\| \gd \|} \left( \dd{w}{\gamma}\transpose \dd{\gamma}{t} \right).
\end{align}
Multiplying both sides by $\| \gd \| / w$, we obtain
\begin{align}
0 &= \|\gd\|^2 \frac{1}{w} \dd{w}{\gamma} - \left( I - \frac{\gd \gd\transpose}{\| \gd \|^2} \right) \gdd  - \gd \frac{1}{w}\dd{w}{\gamma}\transpose \gd\\
&= \|\gd\|^2 \nabla \log w - \left( I - \gdh\gdh\transpose \right)\gdd - \left\langle \nabla \log w, \gd \right\rangle \gd,
\end{align}
where we use that $\nabla \log w = \frac{1}{w} \dd{w}{\gamma}$.
Rearranging, we obtain
\begin{align}
\underbrace{
\left( I - \gdh\gdh\transpose \right)\frac{\gdd}{\|\gd\|^2}
}_{\text{$\perp \gd$}}
&=
\underbrace{
\left( I - \gdh\gdh\transpose \right) \nabla \log w(\gamma)
}_{\text{$\perp \gd$}},
\end{align}
where the unit velocity is given by $\gdh = \gd / \|\gd\|$. In other words, we obtain a relationship between quantities that are both perpendicular to the velocity, one a component of the scaled acceleration and the other a component of the gradient of the log weight.

For a constant speed parameterization, we observe that acceleration in the direction of the path must be zero, and so
$
( I - \gdh\gdh\transpose ) \gdd = \gdd
$.
We therefore obtain
\begin{align}
\gdd
&=
\|\gd\|^2 \left( I - \gdh\gdh\transpose \right) \nabla \log w(\gamma).
\end{align}

\paragraph{Functional derivative.}
This second-order ODE expresses the relationship at optimality, \ie, given an initial position and velocity we can obtain the associated optimal path. However, we can also derive the functional derivative $\ddelta{S}{\gamma}$ of the path length functional $S$ by approximating the curve by a polygonal line with $n$ segments, as $n$ grows arbitrarily large. We obtain, for any (potentially sub-optimal) path $\gamma$,
\begin{align}
\ddelta{S}{\gamma} &= \frac{w(\gamma)}{\|\gd\|}\left( I - \gdh\gdh\transpose \right) \left( \nabla \log w(\gamma) - \frac{\gdd}{\|\gd\|^2} \right),
\label{eq:supp_w_derivative}
\end{align}
or, for a constant speed parameterization,
\begin{align}
\ddelta{S}{\gamma} &= \frac{w(\gamma)}{\|\gd\|} \left( \left( I - \gdh\gdh\transpose \right) \nabla \log w(\gamma) - \frac{\gdd}{\|\gd\|^2} \right). 
\label{eq:supp_w_derivative_constant_speed}
\end{align}

\paragraph{High-probability geodesics.}
In our case of interest, the weight is inversely proportional to the probability density, that is,
\begin{equation}
w(\gamma) = p(\gamma)^{-1}
\text{\; and \;}
\nabla \log w(\gamma) = -\nabla \log p(\gamma),
\end{equation}
giving us, for a constant speed parameterization of the path, the following second-order ODE expressing the optimality condition
\begin{align}
\gdd + \|\gd\|^2 \left( I - \gdh\gdh\transpose \right) \nabla \log p(\gamma) &= 0
\label{eq:supp_w_ode}
\end{align}
and the functional derivative
\begin{align}
\ddelta{S}{\gamma} &= \frac{-1}{p(\gamma)\|\gd\|} \left( \left( I - \gdh\gdh\transpose \right)  \nabla \log p(\gamma) + \frac{\gdd}{\|\gd\|^2} \right).
\label{eq:supp_derivative}
\end{align}

\section{Further Implementation Details}
\label{sec:app_hyperparameters}
For the score function $\phi$ in \cref{eq:score_distillation},  we use a uniform weight function $w(\tau)=1$, and the output is normalized by $1+\sigma$.  
For the negative text prompt, we used ``A doubling image, unrealistic, artifacts, distortions, unnatural blending, ghosting effects, overlapping edges, harsh transitions, motion blur, poor resolution, low detail''
for all the experiments. 
In the inference process of image interpolation, we applied the same perceptually-uniform sampling strategy as \citet{zhang2024diffmorpher} to produce an image sequence with a more homogeneous transition rate, using histogram equalization. 
In both the deterministic DDIM forward (DDIM-F) and backward (DDIM-B) processes on BVP and IVP, we set the classifier-free guidance scale (CFG) to $1$ and use the same positive conditional embedding as the one used in $\phi$ (text-inverted). 
After optimizing each point $x$ in the path, we project the point back to the sphere by scaling the norm of $x$ to the radius of the sphere.
For IVP, we aim to generate an initial velocity that points towards the distribution of the target prompt. 
Given a source and target text embedding $z_0$ and $z_{1}$, we compute a pseudo target $x_\text{tgt}$ by optimizing the initial latent vector $x_0$ using the score function $\phi(x|mz_{1}+(1-m)z_0, \tau)$ with $m=0.8$, learning rate as 1 and number of iteration as 300. 
Then the initial velocity is set as $x_\text{tgt}-x_0$ projected to the tangent space of the sphere. 

\section{Further Details on Evaluation Metrics}
\label{sec:app_eva}

For the TOPIQ score, we weight it to emphasize the quality of the middle frames of the generated sequence, as they tend to be farther from the source images and more indicative of the overall perceptual quality. Instead of a simple average, we compute a weighted TOPIQ score as:
\begin{align}
    \text{TOPIQ}(\{I_{\lambda}\}_{\lambda\in[0,1]}) =  \frac{\sum_{\lambda}  w(\lambda) \text{TOPIQ}(I_\lambda)}{\sum_{\lambda} w(\lambda)} , 
\end{align}
where $w(\lambda)=\lambda$ for $\lambda \leq 0.5$ and $w(\lambda)=1-\lambda$ for $\lambda>0.5$. 

\section{Additional Ablation Study on the Conditioning Signal}
\label{sec:app_cond}
We compare two types of conditioning signals—constant versus linearly varying along the path as discussed in \cref{sec:method_cond}.
For constant conditioning, the text embedding is initialized as $p_0 + p_1$ and then fine-tuned using text inversion.
For time-linear conditioning, we apply text inversion separately to both prompts of the image pair and interpolate their embeddings as $z_t = (1 - t) z_0 + t z_1$, where we use the shorthand $z_t = \zeta(t)$ in this section.
As shown in \cref{tab:condition-path}, constant conditioning results in a distribution closer to the input images (as measured by FID), while the time-linear conditioning yields higher image quality (TOPIQ) and a more homogeneous transition rate (PDV).
The main paper reports results using time-linear conditioning (see \cref{tab-main}).

\begin{table}[!t]
    \centering
    \small
    \setlength{\tabcolsep}{4pt} 
    \renewcommand{\arraystretch}{1.2} 
    \caption{
        Ablation study on the validation dataset that ablates the time-dependence of the conditioning signal ($z_t$) and the geodesic optimization.
        \label{tab:condition-path}
    }
    \label{app_tab-morph_ablation}
    \begin{tabularx}{\linewidth}{@{}ccCCCC@{} } 
    \toprule
    $z_t$ 
    & Opt.
    & FID$\downarrow$
    & PPL$\downarrow$
    & PDV$\downarrow$
    & TOPIQ$\uparrow$ \\
    \midrule
     &  & \textbf{45.30} & \textbf{0.853} & 0.037 & 0.516 \\
     &  $\checkmark$ & 55.80 & 0.931 & 0.020 & 0.541  \\
     $\checkmark$ &   & 49.39 & 0.917 & 0.035 & 0.526\\
    $\checkmark$ & $\checkmark$ & 63.28 & 1.000 & \textbf{0.017} & \textbf{0.553}  \\
    \bottomrule    
    \end{tabularx}
\end{table}

\section{Sensitivity Analysis of the Hyperparameters}
\label{sec:app_anahyper}

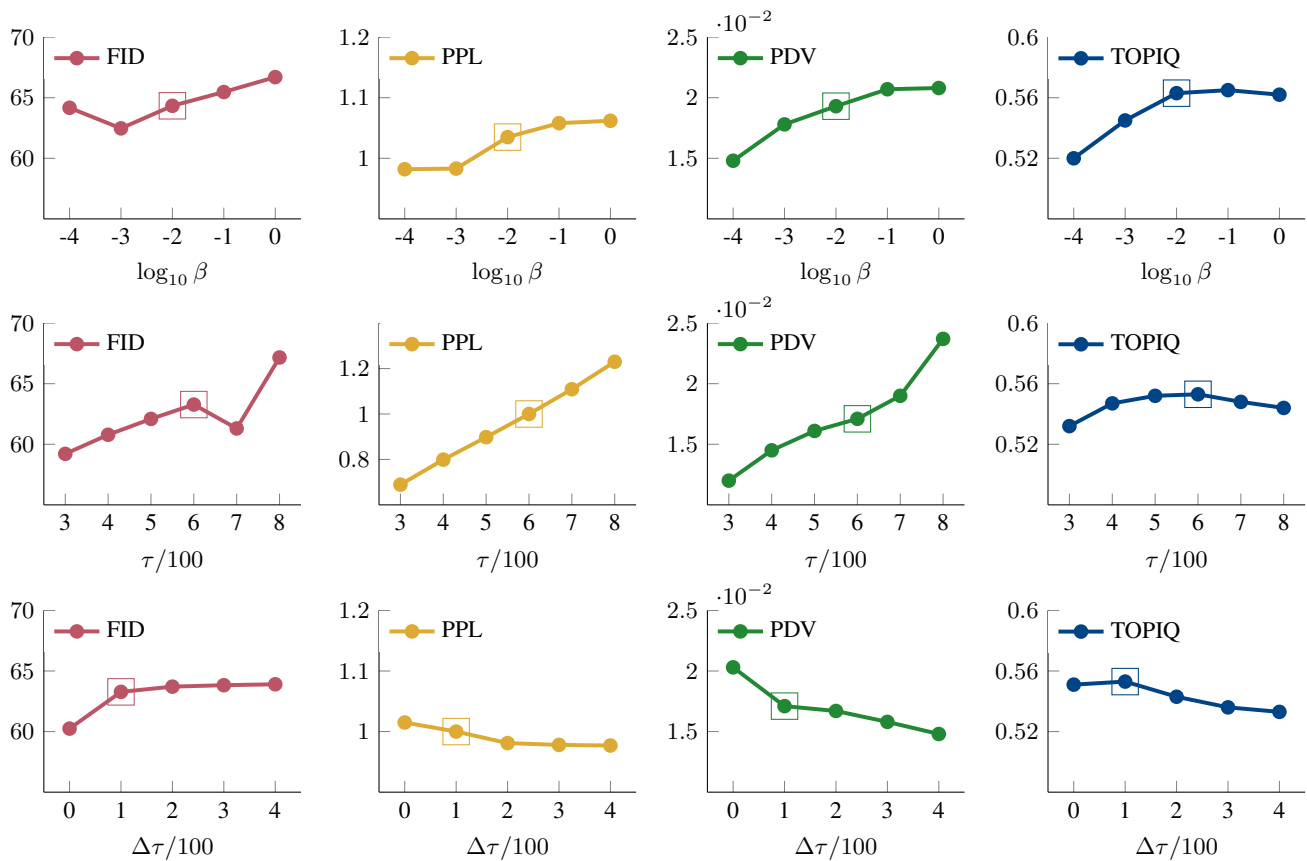
\begin{figure*}[!t]\centering
    \begin{subfigure}{0.25\textwidth}
\begin{tikzpicture}

\definecolor{myyellow}{RGB}{221,170,51} 
\definecolor{myred}{RGB}{187,85,102} 
\definecolor{myblue}{RGB}{0,68,136} 
\definecolor{mygreen}{RGB}{34,136,51} 

\begin{axis}[
width=5cm,
height=4cm,
xmin=0.5,
xmax=5.5,
xtick= {1, 2, 3, 4, 5},
xticklabels={-4, -3, -2, -1, 0}, 
xlabel= {$\log_{10}\beta$},
ymin=55,
ymax=70,
ytick= {60, 65, 70},
ylabel= {},
axis x line*=bottom,
axis y line*=left,
legend entries={ FID },
legend style={draw=none, at={(0.0,1.0)}, anchor=north west},
]

\addplot [color=myred, line width=1.5pt, mark=*] table [row sep=newline]{
1 64.1681652866632
2 62.470967762953
3 64.3325097674945
4 65.469585433779
5 66.700
};
\addplot [color=myred, mark=square, mark size=5pt] table [row sep=newline]{
3 64.3325097674945
};
\end{axis}

\end{tikzpicture}
    \end{subfigure}%
    \begin{subfigure}{0.25\textwidth}
\begin{tikzpicture}

\definecolor{myyellow}{RGB}{221,170,51} 
\definecolor{myred}{RGB}{187,85,102} 
\definecolor{myblue}{RGB}{0,68,136} 
\definecolor{mygreen}{RGB}{34,136,51} 

\begin{axis}[
width=5cm,
height=4cm,
xmin=0.5,
xmax=5.5,
xtick= {1, 2, 3, 4, 5},
xticklabels={-4, -3, -2, -1, 0}, 
xlabel= {$\log_{10}\beta$},
ymin=0.9,
ymax=1.2,
ytick= {1, 1.1, 1.2},
ylabel= {},
axis x line*=bottom,
axis y line*=left,
legend entries={ PPL },
legend style={draw=none, at={(0.0,1.0)}, anchor=north west},
]

\addplot [color=myyellow, line width=1.5pt, mark=*] table [row sep=newline]{
1 0.982
2 0.983
3 1.035
4 1.058
5 1.062
};
\addplot [color=myyellow, mark=square, mark size=5pt] table [row sep=newline]{
3 1.035
};
\end{axis}

\end{tikzpicture}
    \end{subfigure}%
    \begin{subfigure}{0.25\textwidth}
\begin{tikzpicture}

\definecolor{myyellow}{RGB}{221,170,51} 
\definecolor{myred}{RGB}{187,85,102} 
\definecolor{myblue}{RGB}{0,68,136} 
\definecolor{mygreen}{RGB}{34,136,51} 

\begin{axis}[
width=5cm,
height=4cm,
xmin=0.5,
xmax=5.5,
xtick= {1, 2, 3, 4, 5},
xticklabels={-4, -3, -2, -1, 0}, 
xlabel= {$\log_{10}\beta$},
ymin=0.01,
ymax=0.025,
ytick= {0.015, 0.02, 0.025},
ylabel= {},
axis x line*=bottom,
axis y line*=left,
legend entries={ PDV },
legend style={draw=none, at={(0.0,1.0)}, anchor=north west},
]

\addplot [color=mygreen, line width=1.5pt, mark=*] table [row sep=newline]{
1 0.0148
2 0.0178
3 0.0193
4 0.0207
5 0.0208
};
\addplot [color=mygreen, mark=square, mark size=5pt] table [row sep=newline]{
3 0.0193
};
\end{axis}

\end{tikzpicture}
    \end{subfigure}%
    \begin{subfigure}{0.25\textwidth}
\begin{tikzpicture}

\definecolor{myyellow}{RGB}{221,170,51} 
\definecolor{myred}{RGB}{187,85,102} 
\definecolor{myblue}{RGB}{0,68,136} 
\definecolor{mygreen}{RGB}{34,136,51} 

\begin{axis}[
width=5cm,
height=4cm,
xmin=0.5,
xmax=5.5,
xtick= {1, 2, 3, 4, 5},
xticklabels={-4, -3, -2, -1, 0}, 
xlabel= {$\log_{10}\beta$},
ymin=0.48,
ymax=0.6,
ytick= {0.52, 0.56, 0.6},
ylabel= {},
axis x line*=bottom,
axis y line*=left,
legend entries={ TOPIQ },
legend style={draw=none, at={(0.0,1.0)}, anchor=north west},
]

\addplot [color=myblue, line width=1.5pt, mark=*] table [row sep=newline]{
1 0.52
2 0.545
3 0.563
4 0.565
5 0.562
};
\addplot [color=myblue, mark=square, mark size=5pt] table [row sep=newline]{
3 0.563
};
\end{axis}

\end{tikzpicture}
    \end{subfigure} \vfill
      \begin{subfigure}{0.25\textwidth}
\begin{tikzpicture}

\definecolor{myyellow}{RGB}{221,170,51} 
\definecolor{myred}{RGB}{187,85,102} 
\definecolor{myblue}{RGB}{0,68,136} 
\definecolor{mygreen}{RGB}{34,136,51} 

\begin{axis}[
width=5cm,
height=4cm,
xmin=0.5,
xmax=6.5,
xtick= {1, 2, 3, 4, 5, 6},
xticklabels={3, 4, 5, 6, 7, 8}, 
xlabel= {$\tau / 100$},
ymin=55,
ymax=70,
ytick= {60, 65, 70},
ylabel= {},
axis x line*=bottom,
axis y line*=left,
legend entries={ FID },
legend style={draw=none, at={(0.0,1.0)}, anchor=north west},
]

\addplot [color=myred, line width=1.5pt, mark=*] table [row sep=newline]{
1 59.210138951431
2 60.7852463209831
3 62.0940511641115
4 63.2758612414888
5 61.3049320008967
6 67.1633455671619
};
\addplot [color=myred, mark=square, mark size=5pt] table [row sep=newline]{
4 63.2758612414888
};
\end{axis}

\end{tikzpicture}
    \end{subfigure}%
    \begin{subfigure}{0.25\textwidth}
\begin{tikzpicture}

\definecolor{myyellow}{RGB}{221,170,51} 
\definecolor{myred}{RGB}{187,85,102} 
\definecolor{myblue}{RGB}{0,68,136} 
\definecolor{mygreen}{RGB}{34,136,51} 

\begin{axis}[
width=5cm,
height=4cm,
xmin=0.5,
xmax=6.5,
xtick= {1, 2, 3, 4, 5, 6},
xticklabels={3, 4, 5, 6, 7, 8}, 
xlabel= {$\tau / 100$},
ymin=0.6,
ymax=1.4,
ytick= {0.8, 1, 1.2},
ylabel= {},
axis x line*=bottom,
axis y line*=left,
legend entries={ PPL },
legend style={draw=none, at={(0.0,1.0)}, anchor=north west},
]

\addplot [color=myyellow, line width=1.5pt, mark=*] table [row sep=newline]{
1 0.689
2 0.799
3 0.898
4 1.000
5 1.109
6 1.230
};
\addplot [color=myyellow, mark=square, mark size=5pt] table [row sep=newline]{
4 1.000
};
\end{axis}

\end{tikzpicture}
    \end{subfigure}%
    \begin{subfigure}{0.25\textwidth}
\begin{tikzpicture}

\definecolor{myyellow}{RGB}{221,170,51} 
\definecolor{myred}{RGB}{187,85,102} 
\definecolor{myblue}{RGB}{0,68,136} 
\definecolor{mygreen}{RGB}{34,136,51} 

\begin{axis}[
width=5cm,
height=4cm,
xmin=0.5,
xmax=6.5,
xtick= {1, 2, 3, 4, 5, 6},
xticklabels={3, 4, 5, 6, 7, 8}, 
xlabel= {$\tau / 100$},
ymin=0.01,
ymax=0.025,
ytick= {0.015, 0.02, 0.025},
ylabel= {},
axis x line*=bottom,
axis y line*=left,
legend entries={ PDV },
legend style={draw=none, at={(0.0,1.0)}, anchor=north west},
]

\addplot [color=mygreen, line width=1.5pt, mark=*] table [row sep=newline]{
1 0.0120
2 0.0145
3 0.0161
4 0.0171
5 0.0190
6 0.0237
};
\addplot [color=mygreen, mark=square, mark size=5pt] table [row sep=newline]{
4 0.0171
};
\end{axis}

\end{tikzpicture}
    \end{subfigure}%
    \begin{subfigure}{0.25\textwidth}
\begin{tikzpicture}

\definecolor{myyellow}{RGB}{221,170,51} 
\definecolor{myred}{RGB}{187,85,102} 
\definecolor{myblue}{RGB}{0,68,136} 
\definecolor{mygreen}{RGB}{34,136,51} 

\begin{axis}[
width=5cm,
height=4cm,
xmin=0.5,
xmax=6.5,
xtick= {1, 2, 3, 4, 5, 6},
xticklabels={3, 4, 5, 6, 7, 8}, 
xlabel= {$\tau / 100$},
ymin=0.48,
ymax=0.6,
ytick= {0.52, 0.56, 0.6},
ylabel= {},
axis x line*=bottom,
axis y line*=left,
legend entries={ TOPIQ },
legend style={draw=none, at={(0.0,1.0)}, anchor=north west},
]

\addplot [color=myblue, line width=1.5pt, mark=*] table [row sep=newline]{
1 0.532
2 0.547
3 0.552
4 0.553
5 0.548
6 0.544
};
\addplot [color=myblue, mark=square, mark size=5pt] table [row sep=newline]{
4 0.553
};
\end{axis}

\end{tikzpicture}
    \end{subfigure} \vfill
      \begin{subfigure}{0.25\textwidth}
\begin{tikzpicture}

\definecolor{myyellow}{RGB}{221,170,51} 
\definecolor{myred}{RGB}{187,85,102} 
\definecolor{myblue}{RGB}{0,68,136} 
\definecolor{mygreen}{RGB}{34,136,51} 

\begin{axis}[
width=5cm,
height=4cm,
xmin=0.5,
xmax=5.5,
xtick= {1, 2, 3, 4, 5},
xticklabels={0, 1, 2, 3, 4}, 
xlabel= {$ \Delta \tau / 100$},
ymin=55,
ymax=70,
ytick= {60, 65, 70},
ylabel= {},
axis x line*=bottom,
axis y line*=left,
legend entries={ FID },
legend style={draw=none, at={(0.0,1.0)}, anchor=north west},
]

\addplot [color=myred, line width=1.5pt, mark=*] table [row sep=newline]{
1 60.246
2 63.276
3 63.709
4 63.828
5 63.902
};
\addplot [color=myred, mark=square, mark size=5pt] table [row sep=newline]{
2 63.276
};
\end{axis}

\end{tikzpicture}
    \end{subfigure}%
    \begin{subfigure}{0.25\textwidth}
\begin{tikzpicture}

\definecolor{myyellow}{RGB}{221,170,51} 
\definecolor{myred}{RGB}{187,85,102} 
\definecolor{myblue}{RGB}{0,68,136} 
\definecolor{mygreen}{RGB}{34,136,51} 

\begin{axis}[
width=5cm,
height=4cm,
xmin=0.5,
xmax=5.5,
xtick= {1, 2, 3, 4, 5},
xticklabels={0, 1, 2, 3, 4}, 
xlabel= {$\Delta \tau / 100$},
ymin=0.9,
ymax=1.2,
ytick= {1, 1.1, 1.2},
ylabel= {},
axis x line*=bottom,
axis y line*=left,
legend entries={ PPL },
legend style={draw=none, at={(0.0,1.0)}, anchor=north west},
]

\addplot [color=myyellow, line width=1.5pt, mark=*] table [row sep=newline]{
1 1.015
2 1.000
3 0.981
4 0.978
5 0.977
};
\addplot [color=myyellow, mark=square, mark size=5pt] table [row sep=newline]{
2 1.000
};
\end{axis}

\end{tikzpicture}
    \end{subfigure}%
    \begin{subfigure}{0.25\textwidth}
\begin{tikzpicture}

\definecolor{myyellow}{RGB}{221,170,51} 
\definecolor{myred}{RGB}{187,85,102} 
\definecolor{myblue}{RGB}{0,68,136} 
\definecolor{mygreen}{RGB}{34,136,51} 

\begin{axis}[
width=5cm,
height=4cm,
xmin=0.5,
xmax=5.5,
xtick= {1, 2, 3, 4, 5},
xticklabels={0, 1, 2, 3, 4}, 
xlabel= {$\Delta \tau / 100$},
ymin=0.01,
ymax=0.025,
ytick= {0.015, 0.02, 0.025},
ylabel= {},
axis x line*=bottom,
axis y line*=left,
legend entries={ PDV },
legend style={draw=none, at={(0.0,1.0)}, anchor=north west},
]

\addplot [color=mygreen, line width=1.5pt, mark=*] table [row sep=newline]{
1 0.0203
2 0.0171
3 0.0167
4 0.0158
5 0.0148
};
\addplot [color=mygreen, mark=square, mark size=5pt] table [row sep=newline]{
2 0.0171
};
\end{axis}

\end{tikzpicture}
    \end{subfigure}%
    \begin{subfigure}{0.25\textwidth}
\begin{tikzpicture}

\definecolor{myyellow}{RGB}{221,170,51} 
\definecolor{myred}{RGB}{187,85,102} 
\definecolor{myblue}{RGB}{0,68,136} 
\definecolor{mygreen}{RGB}{34,136,51} 

\begin{axis}[
width=5cm,
height=4cm,
xmin=0.5,
xmax=5.5,
xtick= {1, 2, 3, 4, 5},
xticklabels={0, 1, 2, 3, 4}, 
xlabel= {$\Delta \tau / 100$},
ymin=0.48,
ymax=0.6,
ytick= {0.52, 0.56, 0.6},
ylabel= {},
axis x line*=bottom,
axis y line*=left,
legend entries={ TOPIQ },
legend style={draw=none, at={(0.0,1.0)}, anchor=north west},
]

\addplot [color=myblue, line width=1.5pt, mark=*] table [row sep=newline]{
1 0.551
2 0.553
3 0.543
4 0.536
5 0.533
};
\addplot [color=myblue, mark=square, mark size=5pt] table [row sep=newline]{
2 0.553
};
\end{axis}

\end{tikzpicture}
    \end{subfigure}
    \caption{
        The quantitative analysis of selecting hyperparameters $\beta, \tau, \Delta \tau$.
        The default settings are $\beta=0.002, \tau=0.6, \Delta \tau=100$, which are marked in the plots as squares.
	}
    \label{fig:app_ana}
\end{figure*}

We analyze several key hyperparameters of our method, as illustrated in \cref{fig:app_ana}.
The parameter $\beta$ controls the trade-off between path directness and alignment with high-probability regions. 
A larger $\beta$ encourages the path to move toward regions of higher probability density at the expense of directness, while a smaller $\beta$ keeps the path more direct.
The diffusion timestep $\tau$ influences the level of detail in the generated images. 
A higher $\tau$ tends to morph the high-level image features but may result in a loss of fine details, whereas a very low $\tau$ can degrade image quality due to insufficient denoising.
A properly chosen sampling range $\Delta \tau$ can help the path escape local minima compared to using a zero range. However, if $\Delta \tau$ is too large, the guidance signal gets weaker, resulting in smoother paths but lower FID and TOPIQ scores. 
These parameters exhibit interpretable behavior, allowing users to make choices based on their specific needs.

\section{Further Qualitative Results}
\label{sec:app_qual}

\begin{figure*}[!t]\raggedright
     \begin{subfigure}[]{0.97\linewidth}\raggedright
        \includegraphics[width=\linewidth]{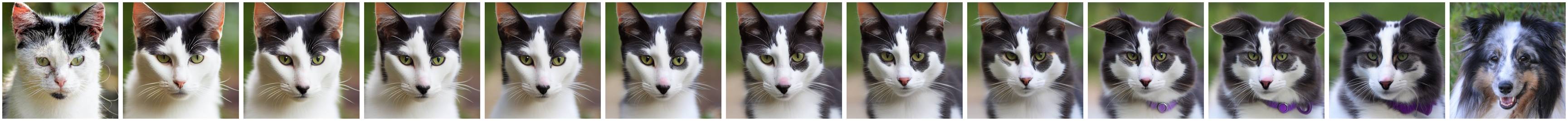}
    \end{subfigure}
    \raisebox{-0.5\height}{\rotatebox{90}{\scriptsize NoiseD.}} \vfill

     \begin{subfigure}[]{0.97\linewidth}\raggedright
        \includegraphics[width=\linewidth]{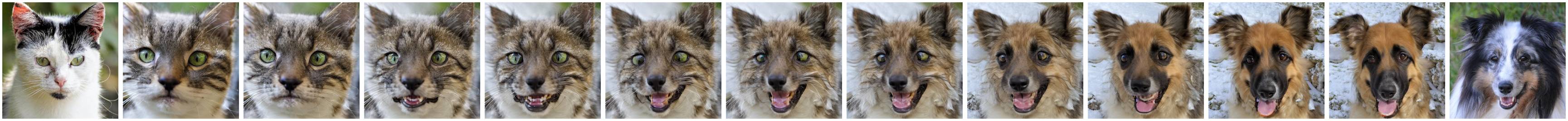}
    \end{subfigure}
    \raisebox{-0.5\height}{\rotatebox{90}{\scriptsize AID}} \vfill
    
    \begin{subfigure}[]{0.97\linewidth}\raggedright
        \includegraphics[width=\linewidth]{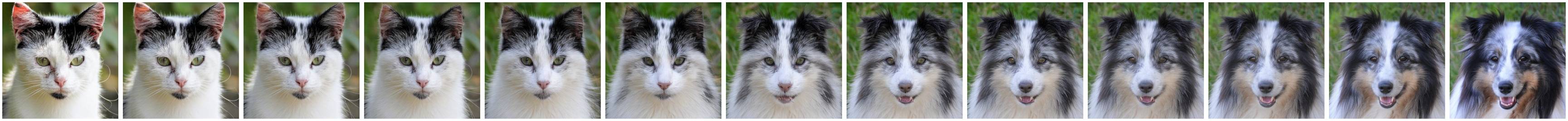}
    \end{subfigure}
    \raisebox{-0.5\height}{\rotatebox{90}{\scriptsize DiffM.}} \vfill
    
   \begin{subfigure}[]{0.97\linewidth}\raggedright
        \includegraphics[width=\linewidth]{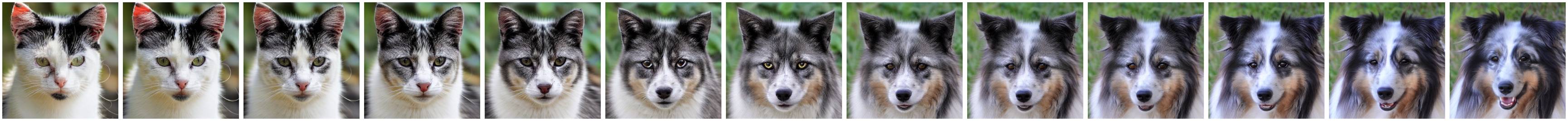}
    \end{subfigure}
    \raisebox{-0.5\height}{\rotatebox{90}{\scriptsize IMPUS}} \vfill

    \begin{subfigure}[]{0.97\linewidth}\raggedright
        \includegraphics[width=\linewidth]{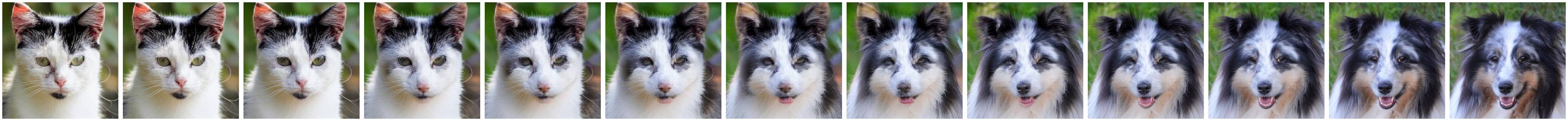}
    \end{subfigure}
    \raisebox{-0.5\height}{\rotatebox{90}{\scriptsize SmoothD.}} \vfill

    \begin{subfigure}[]{0.97\linewidth}\raggedright
        \includegraphics[width=\linewidth]{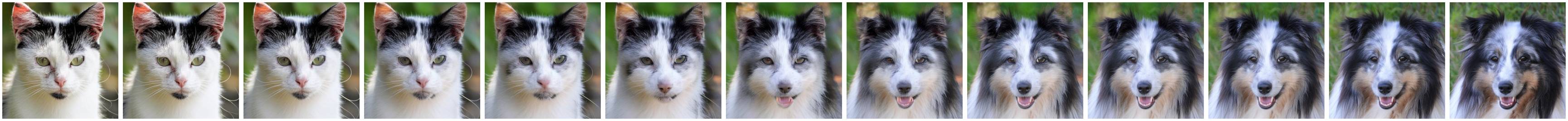}
    \end{subfigure}
    \raisebox{-0.5\height}{\rotatebox{90}{\scriptsize Ours w/o opt.}} \vfill
    
    \begin{subfigure}[]{0.97\linewidth}\raggedright
        \includegraphics[width=\linewidth]{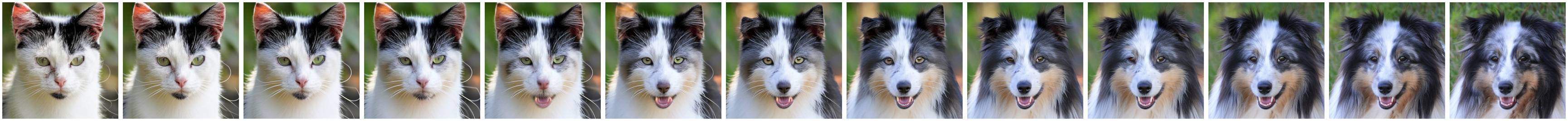}
    \end{subfigure}
    \raisebox{-0.5\height}{\rotatebox{90}{\scriptsize Ours}}
    \caption{
        Qualitative image interpolation results, comparing all methods. 
	}
    \label{fig:qualitative_1}
\end{figure*}

\begin{figure*}[!t]\raggedright
     \begin{subfigure}[]{0.97\linewidth}\raggedright
        \includegraphics[width=\linewidth]{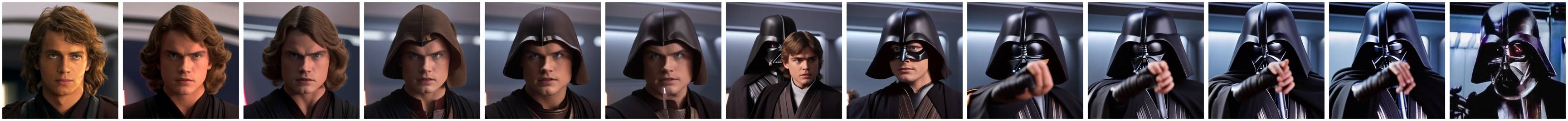}
    \end{subfigure}
    \raisebox{-0.5\height}{\rotatebox{90}{\scriptsize NoiseD.}} \vfill

     \begin{subfigure}[]{0.97\linewidth}\raggedright
        \includegraphics[width=\linewidth]{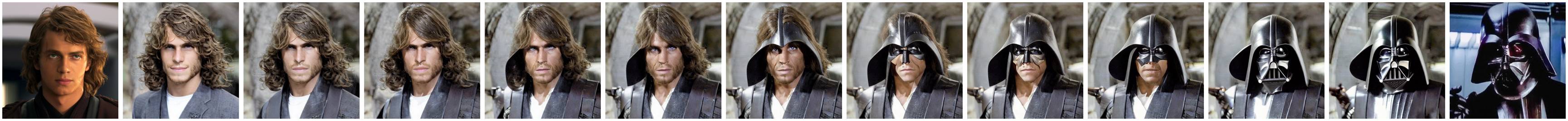}
    \end{subfigure}
    \raisebox{-0.5\height}{\rotatebox{90}{\scriptsize AID}} \vfill
    
    \begin{subfigure}[]{0.97\linewidth}\raggedright
        \includegraphics[width=\linewidth]{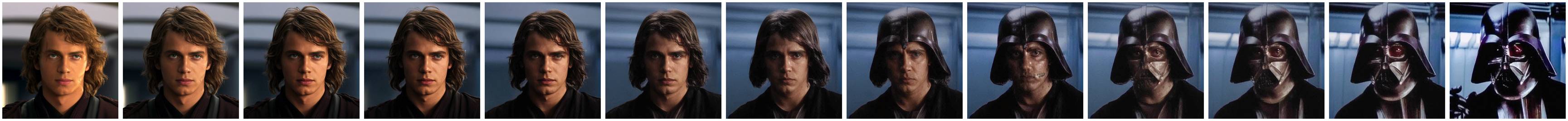}
    \end{subfigure}
    \raisebox{-0.5\height}{\rotatebox{90}{\scriptsize DiffM.}} \vfill
    
   \begin{subfigure}[]{0.97\linewidth}\raggedright
        \includegraphics[width=\linewidth]{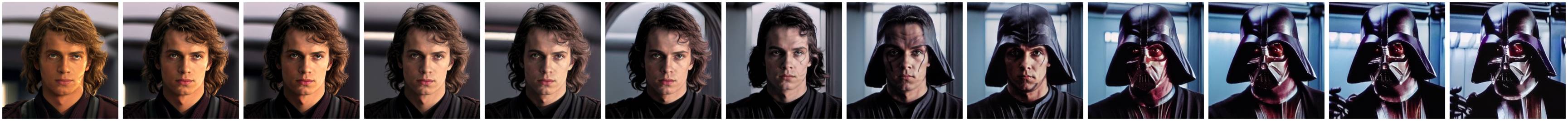}
    \end{subfigure}
    \raisebox{-0.5\height}{\rotatebox{90}{\scriptsize IMPUS}} \vfill

    \begin{subfigure}[]{0.97\linewidth}\raggedright
        \includegraphics[width=\linewidth]{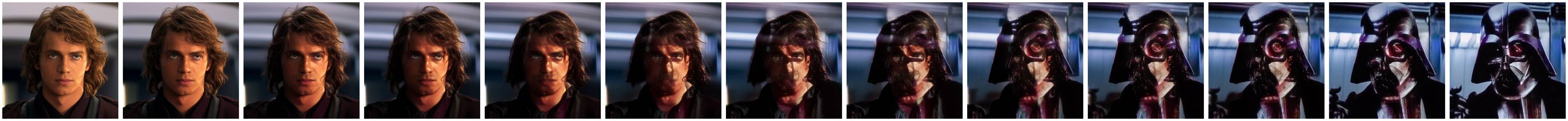}
    \end{subfigure}
    \raisebox{-0.5\height}{\rotatebox{90}{\scriptsize SmoothD.}} \vfill

    \begin{subfigure}[]{0.97\linewidth}\raggedright
        \includegraphics[width=\linewidth]{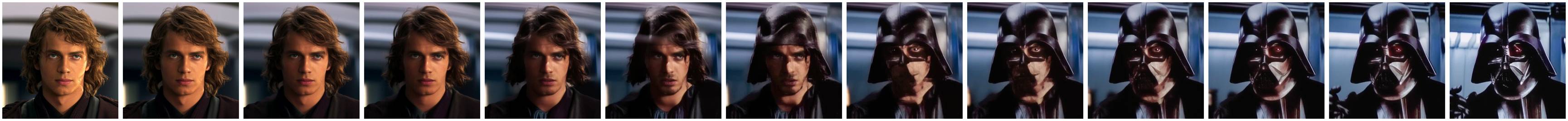}
    \end{subfigure}
    \raisebox{-0.5\height}{\rotatebox{90}{\scriptsize Ours w/o opt.}} \vfill
    
    \begin{subfigure}[]{0.97\linewidth}\raggedright
        \includegraphics[width=\linewidth]{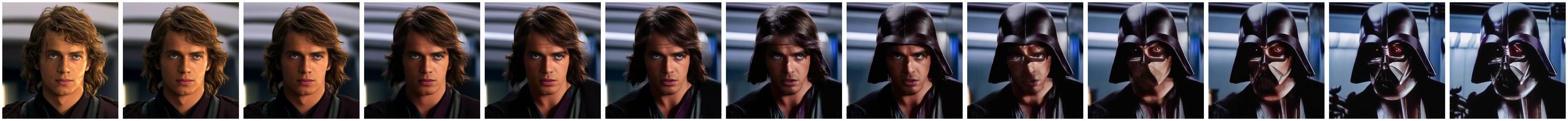}
    \end{subfigure}
    \raisebox{-0.5\height}{\rotatebox{90}{\scriptsize Ours}}
    \caption{
        Qualitative image interpolation results, comparing all methods. 
	}
    \label{fig:qualitative_2}
\end{figure*}

\begin{figure*}[!t]\raggedright
     \begin{subfigure}[]{0.97\linewidth}\raggedright
        \includegraphics[width=\linewidth]{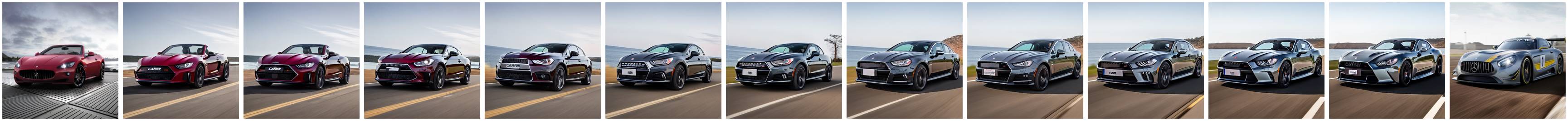}
    \end{subfigure}
    \raisebox{-0.5\height}{\rotatebox{90}{\scriptsize NoiseD.}} \vfill

     \begin{subfigure}[]{0.97\linewidth}\raggedright
        \includegraphics[width=\linewidth]{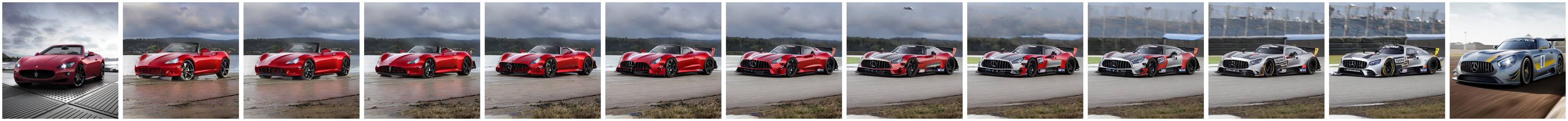}
    \end{subfigure}
    \raisebox{-0.5\height}{\rotatebox{90}{\scriptsize AID}} \vfill
    
    \begin{subfigure}[]{0.97\linewidth}\raggedright
        \includegraphics[width=\linewidth]{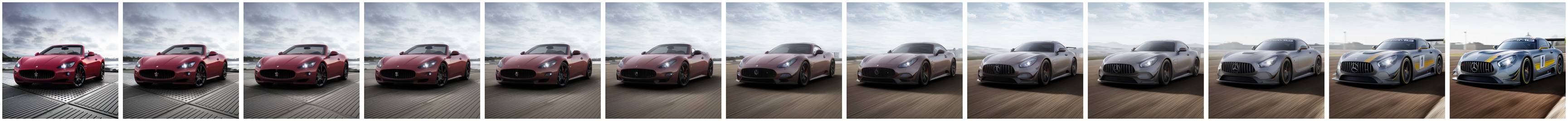}
    \end{subfigure}
    \raisebox{-0.5\height}{\rotatebox{90}{\scriptsize DiffM.}} \vfill
    
   \begin{subfigure}[]{0.97\linewidth}\raggedright
        \includegraphics[width=\linewidth]{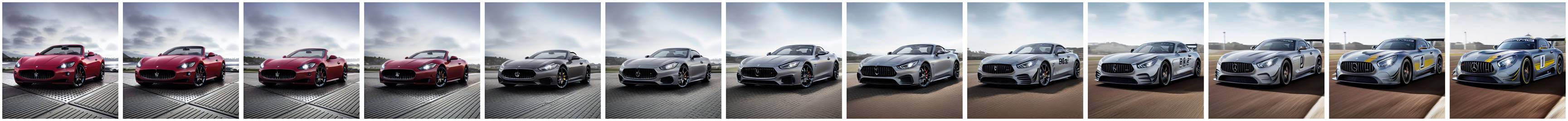}
    \end{subfigure}
    \raisebox{-0.5\height}{\rotatebox{90}{\scriptsize IMPUS}} \vfill

    \begin{subfigure}[]{0.97\linewidth}\raggedright
        \includegraphics[width=\linewidth]{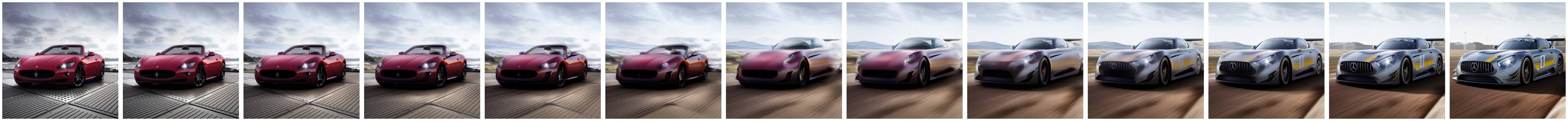}
    \end{subfigure}
    \raisebox{-0.5\height}{\rotatebox{90}{\scriptsize SmoothD.}} \vfill

    \begin{subfigure}[]{0.97\linewidth}\raggedright
        \includegraphics[width=\linewidth]{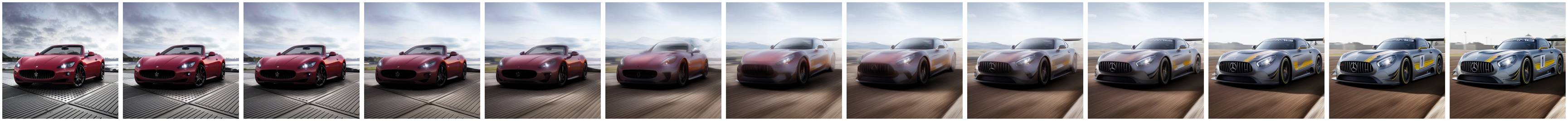}
    \end{subfigure}
    \raisebox{-0.5\height}{\rotatebox{90}{\scriptsize Ours w/o opt.}} \vfill
    
    \begin{subfigure}[]{0.97\linewidth}\raggedright
        \includegraphics[width=\linewidth]{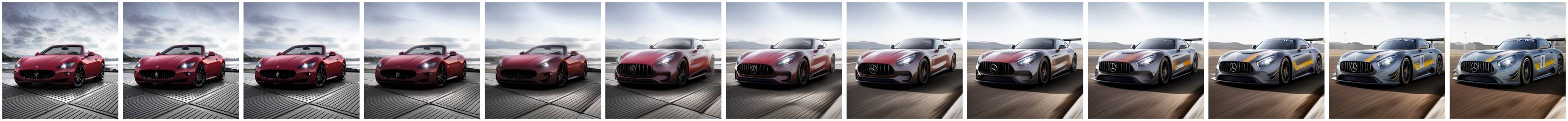}
    \end{subfigure}
    \raisebox{-0.5\height}{\rotatebox{90}{\scriptsize Ours}}
    \caption{
        Qualitative image interpolation results, comparing all methods. 
	}
    \label{fig:qualitative_3}
\end{figure*}

\begin{figure*}[!t]\raggedright
     \begin{subfigure}[]{0.97\linewidth}\raggedright
        \includegraphics[width=\linewidth]{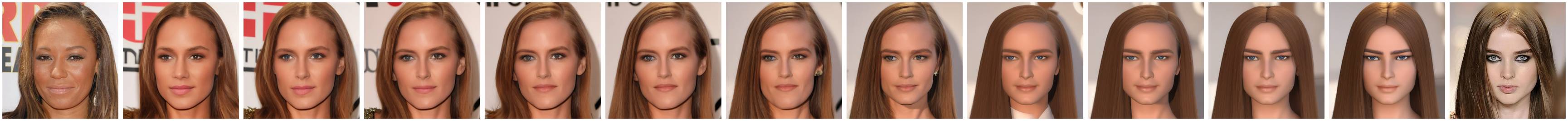}
    \end{subfigure}
    \raisebox{-0.5\height}{\rotatebox{90}{\scriptsize NoiseD.}} \vfill

     \begin{subfigure}[]{0.97\linewidth}\raggedright
        \includegraphics[width=\linewidth]{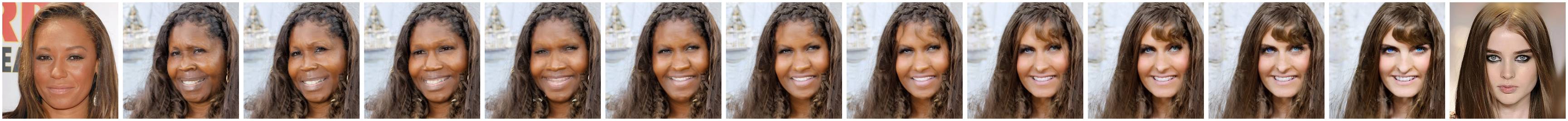}
    \end{subfigure}
    \raisebox{-0.5\height}{\rotatebox{90}{\scriptsize AID}} \vfill
    
    \begin{subfigure}[]{0.97\linewidth}\raggedright
        \includegraphics[width=\linewidth]{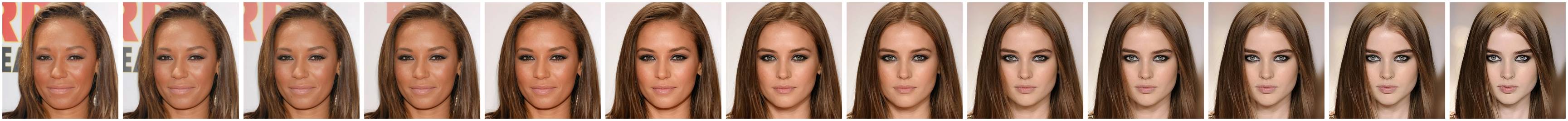}
    \end{subfigure}
    \raisebox{-0.5\height}{\rotatebox{90}{\scriptsize DiffM.}} \vfill
    
   \begin{subfigure}[]{0.97\linewidth}\raggedright
        \includegraphics[width=\linewidth]{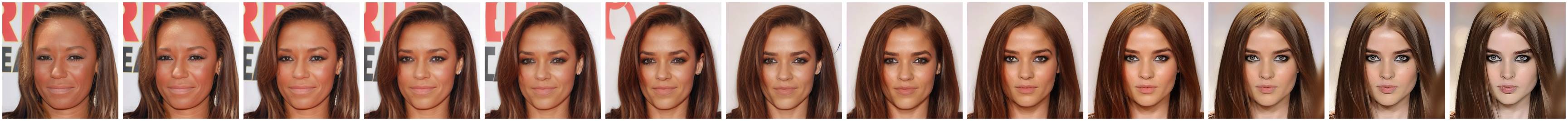}
    \end{subfigure}
    \raisebox{-0.5\height}{\rotatebox{90}{\scriptsize IMPUS}} \vfill

    \begin{subfigure}[]{0.97\linewidth}\raggedright
        \includegraphics[width=\linewidth]{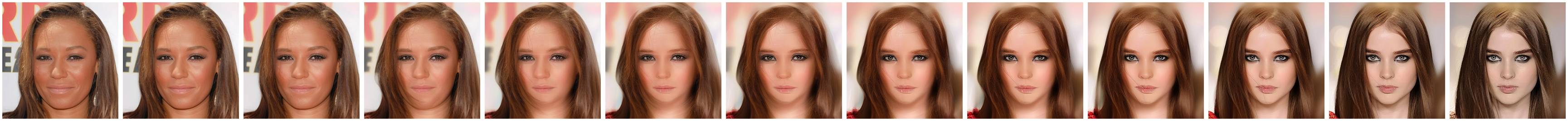}
    \end{subfigure}
    \raisebox{-0.5\height}{\rotatebox{90}{\scriptsize SmoothD.}} \vfill

    \begin{subfigure}[]{0.97\linewidth}\raggedright
        \includegraphics[width=\linewidth]{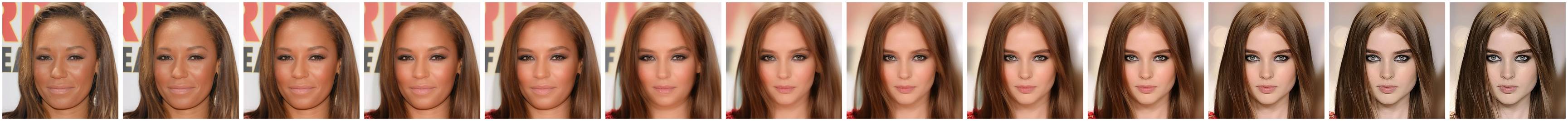}
    \end{subfigure}
    \raisebox{-0.5\height}{\rotatebox{90}{\scriptsize Ours w/o opt.}} \vfill
    
    \begin{subfigure}[]{0.97\linewidth}\raggedright
        \includegraphics[width=\linewidth]{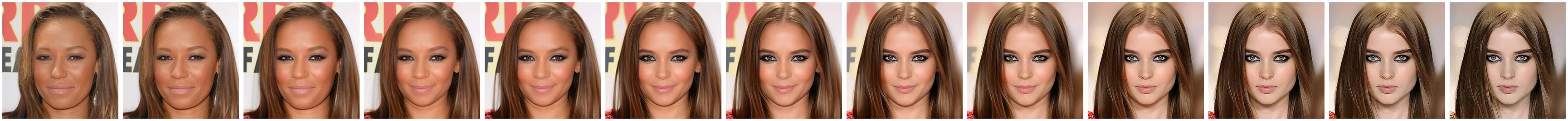}
    \end{subfigure}
    \raisebox{-0.5\height}{\rotatebox{90}{\scriptsize Ours}}
    \caption{
        Qualitative image interpolation results, comparing all methods. 
	}
    \label{fig:qualitative_4}
\end{figure*}

\begin{figure*}[!t]\raggedright
     \begin{subfigure}[]{0.97\linewidth}\raggedright
        \includegraphics[width=\linewidth]{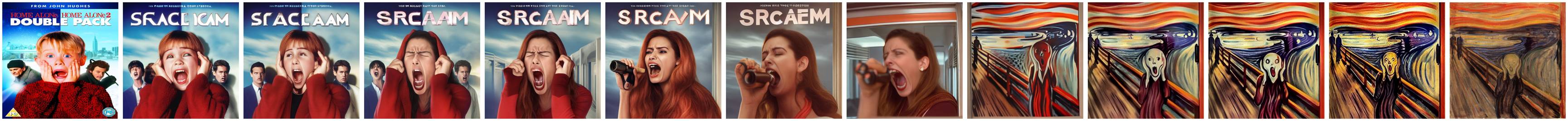}
    \end{subfigure}
    \raisebox{-0.5\height}{\rotatebox{90}{\scriptsize NoiseD.}} \vfill

     \begin{subfigure}[]{0.97\linewidth}\raggedright
        \includegraphics[width=\linewidth]{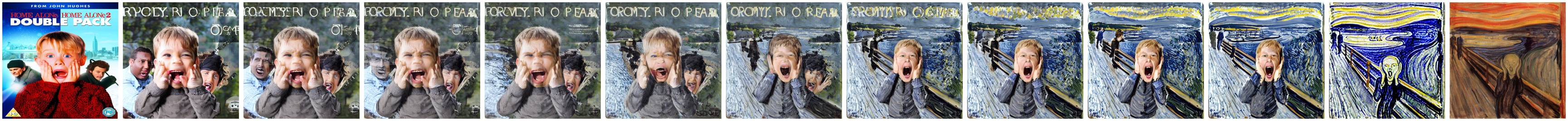}
    \end{subfigure}
    \raisebox{-0.5\height}{\rotatebox{90}{\scriptsize AID}} \vfill
    
    \begin{subfigure}[]{0.97\linewidth}\raggedright
        \includegraphics[width=\linewidth]{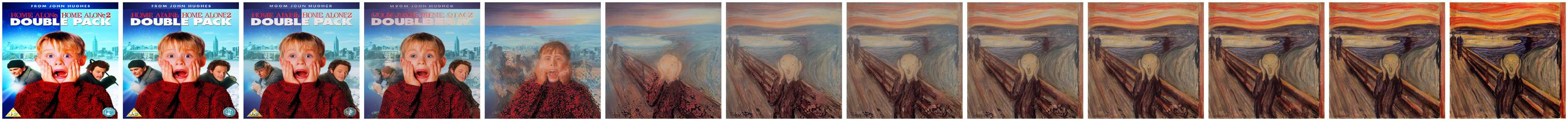}
    \end{subfigure}
    \raisebox{-0.5\height}{\rotatebox{90}{\scriptsize DiffM.}} \vfill
    
   \begin{subfigure}[]{0.97\linewidth}\raggedright
        \includegraphics[width=\linewidth]{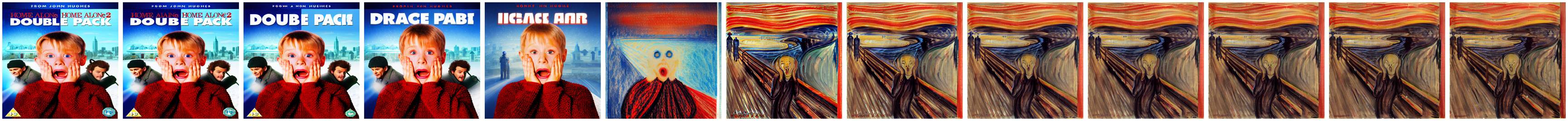}
    \end{subfigure}
    \raisebox{-0.5\height}{\rotatebox{90}{\scriptsize IMPUS}} \vfill

    \begin{subfigure}[]{0.97\linewidth}\raggedright
        \includegraphics[width=\linewidth]{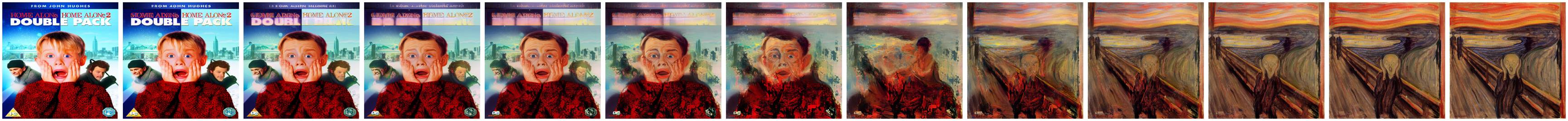}
    \end{subfigure}
    \raisebox{-0.5\height}{\rotatebox{90}{\scriptsize SmoothD.}} \vfill

    \begin{subfigure}[]{0.97\linewidth}\raggedright
        \includegraphics[width=\linewidth]{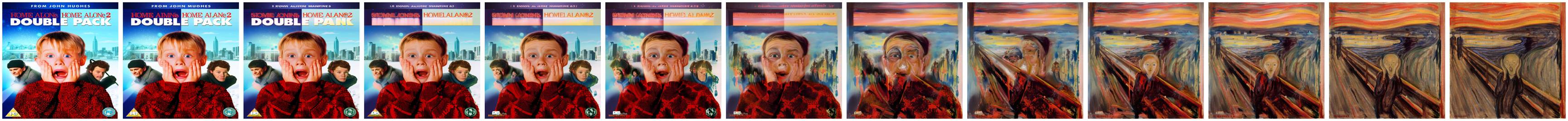}
    \end{subfigure}
    \raisebox{-0.5\height}{\rotatebox{90}{\scriptsize Ours w/o opt.}} \vfill
    
    \begin{subfigure}[]{0.97\linewidth}\raggedright
        \includegraphics[width=\linewidth]{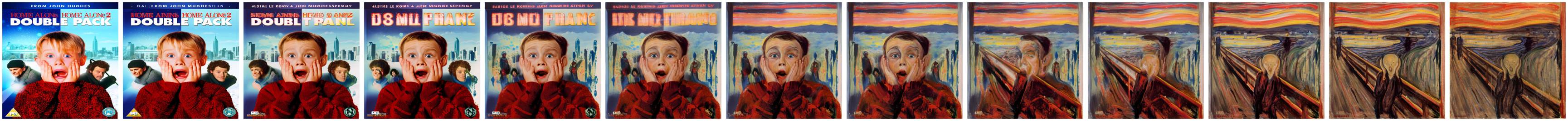}
    \end{subfigure}
    \raisebox{-0.5\height}{\rotatebox{90}{\scriptsize Ours}}
    \caption{
        Qualitative image interpolation results, comparing all methods. 
	}
    \label{fig:qualitative_5}
\end{figure*}

\begin{figure*}[!t]\raggedright
     \begin{subfigure}[]{0.97\linewidth}\raggedright
        \includegraphics[width=\linewidth]{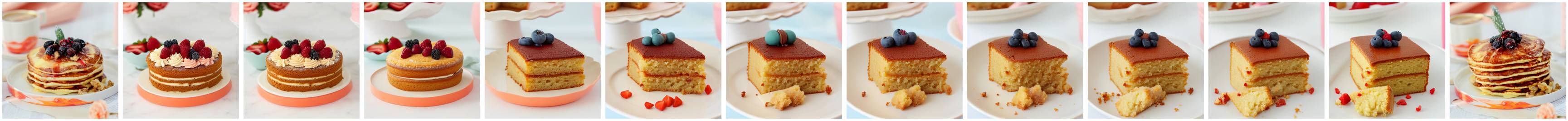}
    \end{subfigure}
    \raisebox{-0.5\height}{\rotatebox{90}{\scriptsize NoiseD.}} \vfill

     \begin{subfigure}[]{0.97\linewidth}\raggedright
        \includegraphics[width=\linewidth]{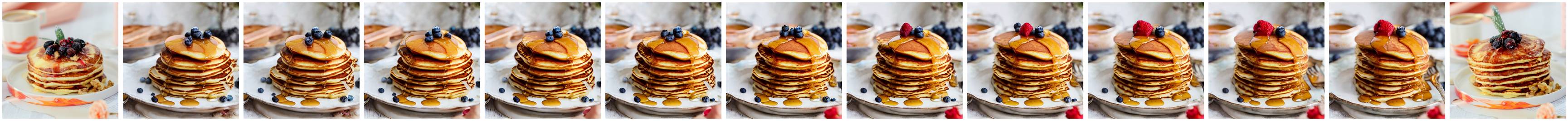}
    \end{subfigure}
    \raisebox{-0.5\height}{\rotatebox{90}{\scriptsize AID}} \vfill
    
    \begin{subfigure}[]{0.97\linewidth}\raggedright
        \includegraphics[width=\linewidth]{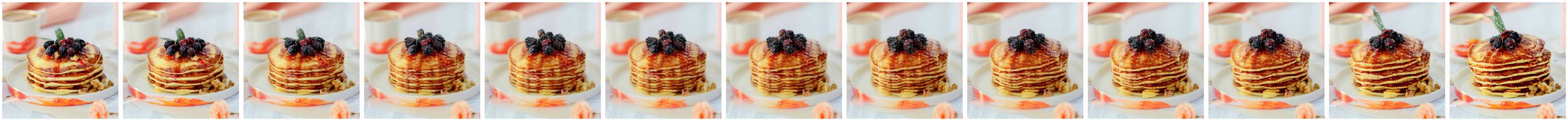}
    \end{subfigure}
    \raisebox{-0.5\height}{\rotatebox{90}{\scriptsize DiffM.}} \vfill
    
   \begin{subfigure}[]{0.97\linewidth}\raggedright
        \includegraphics[width=\linewidth]{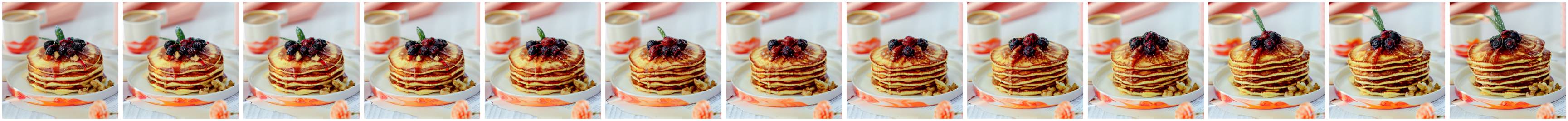}
    \end{subfigure}
    \raisebox{-0.5\height}{\rotatebox{90}{\scriptsize IMPUS}} \vfill

    \begin{subfigure}[]{0.97\linewidth}\raggedright
        \includegraphics[width=\linewidth]{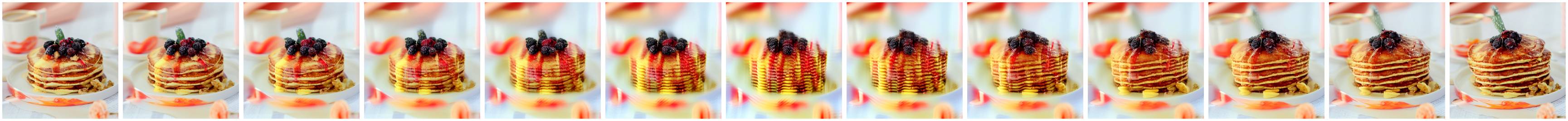}
    \end{subfigure}
    \raisebox{-0.5\height}{\rotatebox{90}{\scriptsize SmoothD.}} \vfill

    \begin{subfigure}[]{0.97\linewidth}\raggedright
        \includegraphics[width=\linewidth]{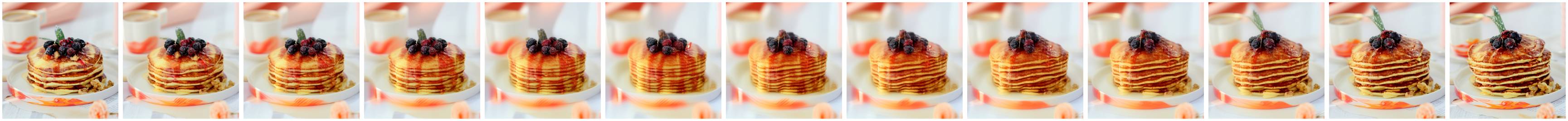}
    \end{subfigure}
    \raisebox{-0.5\height}{\rotatebox{90}{\scriptsize Ours w/o opt.}} \vfill
    
    \begin{subfigure}[]{0.97\linewidth}\raggedright
        \includegraphics[width=\linewidth]{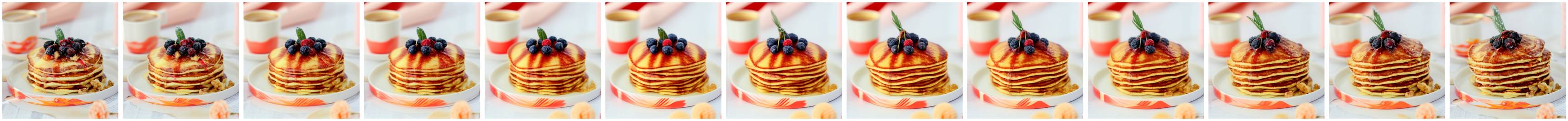}
    \end{subfigure}
    \raisebox{-0.5\height}{\rotatebox{90}{\scriptsize Ours}}
    \caption{
        Qualitative image interpolation results, comparing all methods. 
	}
    \label{fig:qualitative_6}
\end{figure*}

\begin{figure*}[!t]\centering
     \begin{subfigure}[]{\linewidth}\centering
        \includegraphics[width=\linewidth]{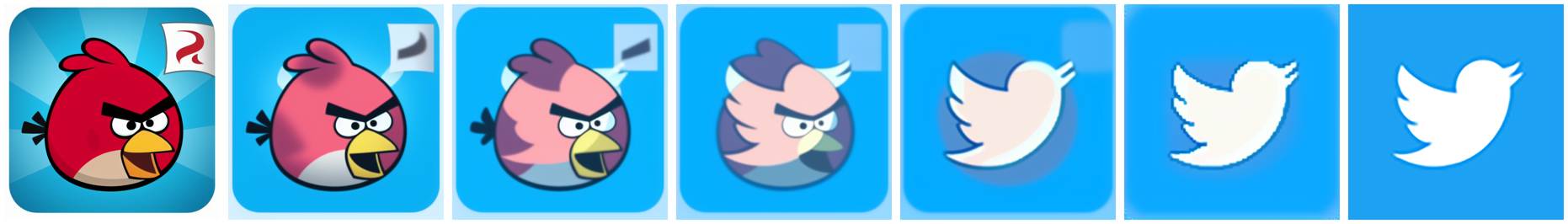}
    \end{subfigure} \vfill
    \begin{subfigure}[]{\linewidth}\centering
        \includegraphics[width=\linewidth]{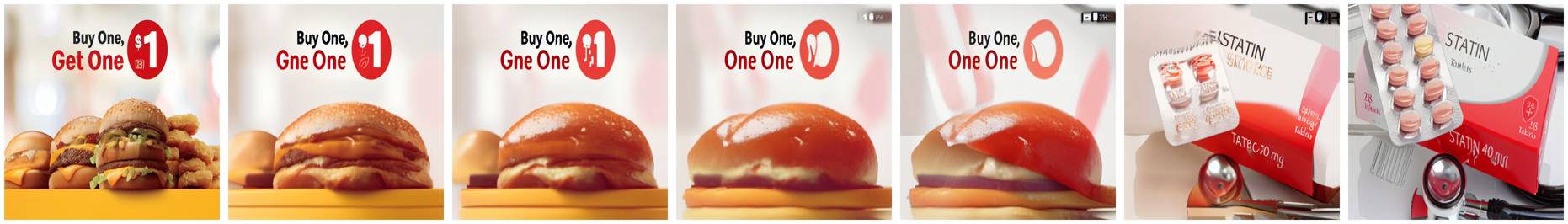}
    \end{subfigure} \vfill
   \begin{subfigure}[]{\linewidth}\centering
        \includegraphics[width=\linewidth]{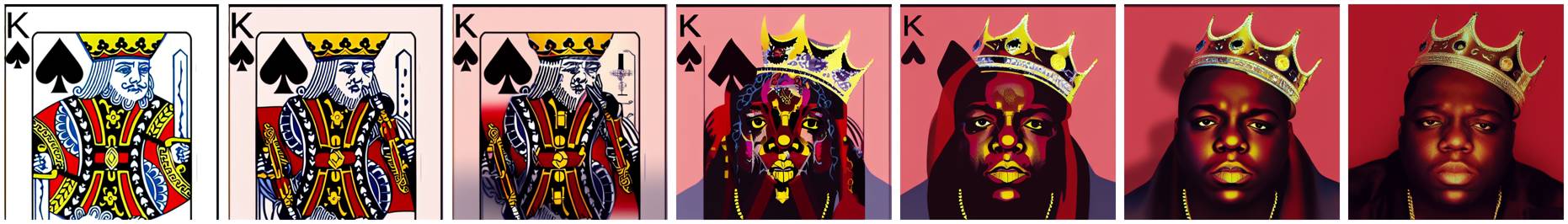}
    \end{subfigure} \vfill
     \begin{subfigure}[]{\linewidth}\centering
        \includegraphics[width=\linewidth]{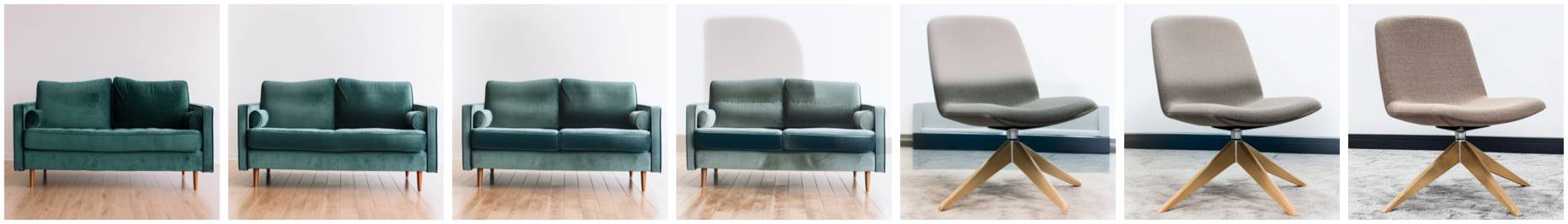}
    \end{subfigure} 
     \begin{subfigure}[]{\linewidth}\centering
        \includegraphics[width=\linewidth]{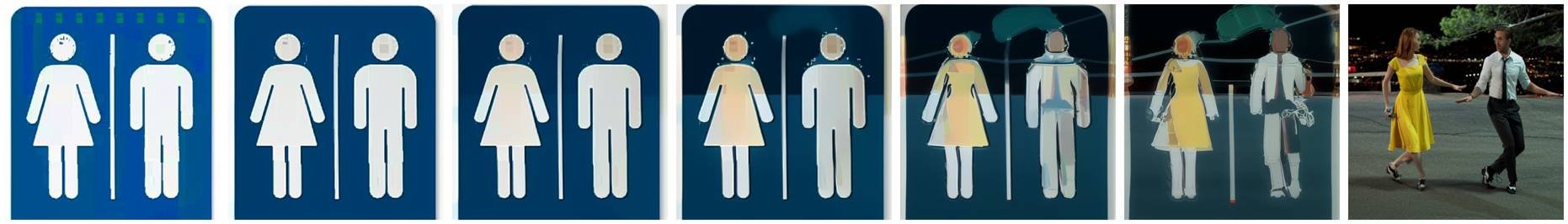}
    \end{subfigure} 
    \caption{
        Image interpolation failure cases.
        Here we show examples with a significant appearance or semantic gap between the image pairs, where the computed geodesic is unable to smoothly connect the two.
	}
    \label{fig:interp_failure_1}
\end{figure*}

\begin{figure*}[!t]\centering
     \begin{subfigure}[]{\linewidth}\centering
        \includegraphics[width=\linewidth]{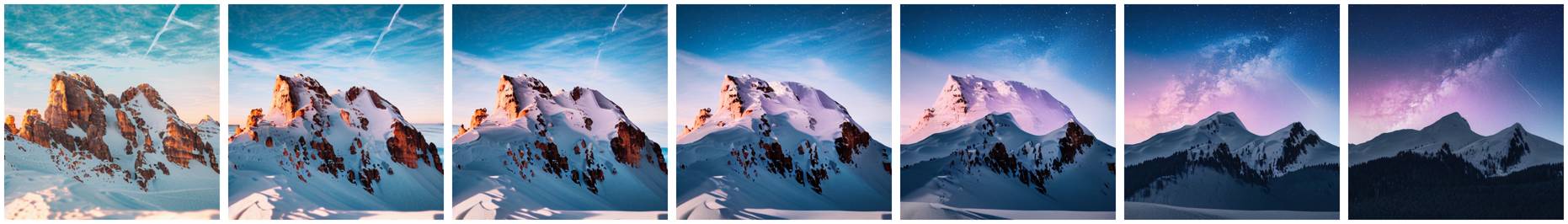}
    \end{subfigure} \vfill
     \begin{subfigure}[]{\linewidth}\centering
        \includegraphics[width=\linewidth]{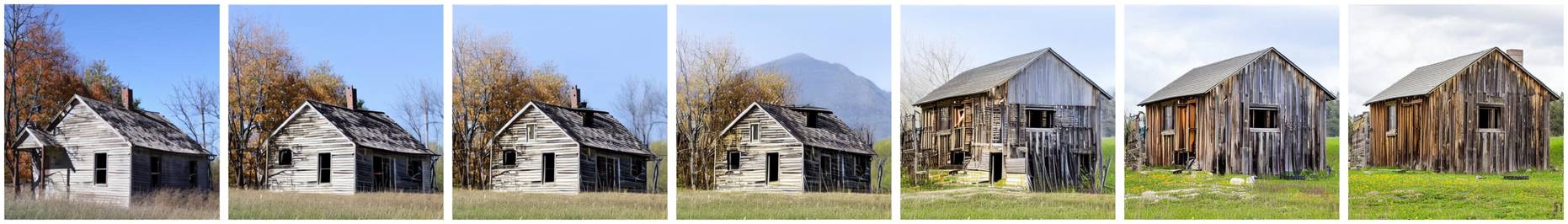}
    \end{subfigure}
    \caption{
        Image interpolation failure cases.
        Here we show how smoothness in image space does not necessarily correspond to smoothness in the projected 3D world.
        For example, in the top row we see a shadow boundary become a ridge line, and in the bottom row we see a mountain become a roof line. 
	}
    \label{fig:interp_failure_2}
\end{figure*}

In this section, we present additional qualitative results and failure cases, as shown in 
\cref{fig:qualitative_1,fig:qualitative_2,fig:qualitative_3,fig:qualitative_4,fig:qualitative_5,fig:qualitative_6,fig:interp_failure_1,fig:interp_failure_2}.

\clearpage

\label{sec:app_exp}

\end{document}